\documentclass[runningheads]{llncs}
\usepackage[T1]{fontenc}
\usepackage{graphicx}
\usepackage{algorithmic}
\usepackage[linesnumbered,ruled]{algorithm2e}
\usepackage{url}
\usepackage{amsmath}
\usepackage{float}
\usepackage{amssymb}
\usepackage{comment}
\usepackage{grffile}
\usepackage{graphicx}
\usepackage{pgf, tikz}
\usetikzlibrary{arrows}

\usepackage{newfloat}
\usepackage{listings}
\usepackage{hyperref}

\title{On Minimum Aerial Photographs for Planar Region Coverage: Hardness and Approximation}
\author{
    Si Wei Feng\inst{1} \and Kartik Mohta\inst{1} \and Kangli Wang\inst{1}
}

\institute{
    Autel Robotics
}
\authorrunning{S. Feng, K. Mohta and K. Wang}
\titlerunning{Minimum Aerial Photographs for Planar Region Coverage}
\usepackage{bibentry}

\begin{document}

\maketitle

\begin{abstract} 
  Aerial photography with drones often requires covering a planar region with a limited number of
  images while maximizing image resolution, equivalently minimizing the footprint size of each photograph.
  We study this task as covering a simple planar polygon with $k$ equal squares or circles of minimum size,
  including the practically relevant variant in which photograph centers must lie inside the region or on its boundary.
  We prove that approximating the minimum square side length is NP-hard within a factor of $1.165$,
  and within a factor of $1.25$ when square centers are restricted to the region;
  together with known hardness for circle coverage, these gaps establish strong intractability for aerial coverage planning.
  We further give a $(2\sqrt{2}+\epsilon)$-approximation algorithm for square coverage via sampling and
  farthest-point clustering under the $L_\infty$ metric, which also applies under the center-location constraints.
  Beyond aerial surveying, the results inform related geometric covering tasks such as facility and sensor placement.
\end{abstract}

\section{Introduction}

Consider the scenario when a drone is tasked to take pictures to capture the 
boundary of some cropland with a gimbal camera attached to it. 
We can change the zoom factor $\lambda$ of the gimbal camera to make the image footprint larger or smaller with the same resolution, 
e.g., 6K resolution ($8000 \times 6000$) on Autel EVO Lite 640T\footnote{\url{https://www.autelrobotics.com/productdetail/evo-lite-enterprise-series/}}.
The gimbal camera usually faces downwards when the drone flies high above the ground to take pictures 
for applications such as security inspection, agricultural analysis, or structure mapping. 
With a greater zoom factor of the gimbal camera, 
the projected footprint on the ground becomes smaller, but the picture taken contains more details 
for the same size of area on the ground. 

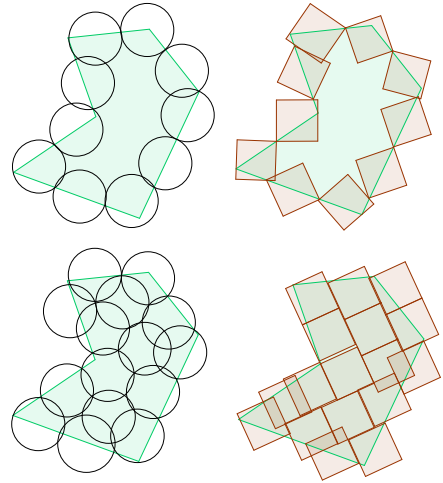
\begin{figure}[h]
\centering
\resizebox{0.9\linewidth}{!}{%
\definecolor{xdxdff}{rgb}{0.49,0.49,1}
\definecolor{uququq}{rgb}{0.25,0.25,0.25}
\definecolor{zzttqq}{rgb}{0.6,0.2,0}
\definecolor{qqccww}{rgb}{0,0.8,0.4}
\definecolor{qqqqff}{rgb}{0,0,1}
\begin{tikzpicture}[line cap=round,line join=round,>=triangle 45,x=1.0cm,y=1.0cm]
\clip(-18.06,-22.1) rectangle (11,8.61);
\fill[color=qqccww,fill=qqccww,fill opacity=0.1] (0.45,3.23) -- (4.68,3.68) -- (7.26,0.42) -- (4.17,-6.15) -- (-2.37,-3.77) -- (1.9,-0.87) -- cycle;
\fill[color=zzttqq,fill=zzttqq,fill opacity=0.1] (0.23,2.99) -- (2.06,1.73) -- (3.32,3.56) -- (1.49,4.82) -- cycle;
\fill[color=zzttqq,fill=zzttqq,fill opacity=0.1] (3.32,3.56) -- (3.95,1.73) -- (5.78,2.36) -- (5.15,4.19) -- cycle;
\fill[color=zzttqq,fill=zzttqq,fill opacity=0.1] (5.78,2.36) -- (5.26,0.39) -- (7.23,-0.13) -- (7.75,1.84) -- cycle;
\fill[color=zzttqq,fill=zzttqq,fill opacity=0.1] (5.16,-0.68) -- (5.81,-2.63) -- (7.77,-1.98) -- (7.11,-0.02) -- cycle;
\fill[color=zzttqq,fill=zzttqq,fill opacity=0.1] (3.86,-3.33) -- (4.53,-5.29) -- (6.5,-4.62) -- (5.82,-2.65) -- cycle;
\fill[color=zzttqq,fill=zzttqq,fill opacity=0.1] (1.9,-5.41) -- (3.24,-6.95) -- (4.79,-5.6) -- (3.44,-4.06) -- cycle;
\fill[color=zzttqq,fill=zzttqq,fill opacity=0.1] (-0.79,-4.34) -- (0.08,-6.23) -- (1.97,-5.36) -- (1.1,-3.47) -- cycle;
\fill[color=zzttqq,fill=zzttqq,fill opacity=0.1] (-2.31,-2.26) -- (-2.38,-4.31) -- (-0.34,-4.38) -- (-0.27,-2.34) -- cycle;
\fill[color=zzttqq,fill=zzttqq,fill opacity=0.1] (-0.27,-2.34) -- (1.88,-2.33) -- (1.88,-0.18) -- (-0.27,-0.19) -- cycle;
\fill[color=zzttqq,fill=zzttqq,fill opacity=0.1] (1.65,-0.14) -- (2.53,1.72) -- (0.67,2.61) -- (-0.22,0.74) -- cycle;
\fill[color=qqccww,fill=qqccww,fill opacity=0.1] (-11.1,3.04) -- (-6.87,3.49) -- (-4.29,0.23) -- (-7.38,-6.34) -- (-13.92,-3.95) -- (-9.65,-1.06) -- cycle;
\fill[color=qqccww,fill=qqccww,fill opacity=0.1] (-11.13,-9.59) -- (-6.91,-9.15) -- (-4.33,-12.4) -- (-7.42,-18.98) -- (-13.96,-16.59) -- (-9.69,-13.69) -- cycle;
\fill[color=qqccww,fill=qqccww,fill opacity=0.1] (0.59,-9.8) -- (4.81,-9.36) -- (7.39,-12.61) -- (4.3,-19.19) -- (-2.24,-16.8) -- (2.03,-13.9) -- cycle;
\fill[color=zzttqq,fill=zzttqq,fill opacity=0.1] (0.21,-9.99) -- (1.08,-11.88) -- (2.97,-11.01) -- (2.1,-9.12) -- cycle;
\fill[color=zzttqq,fill=zzttqq,fill opacity=0.1] (2.41,-9.71) -- (3.28,-11.6) -- (5.17,-10.73) -- (4.3,-8.84) -- cycle;
\fill[color=zzttqq,fill=zzttqq,fill opacity=0.1] (4.53,-9.47) -- (5.4,-11.36) -- (7.29,-10.49) -- (6.42,-8.6) -- cycle;
\fill[color=zzttqq,fill=zzttqq,fill opacity=0.1] (1.08,-11.88) -- (1.95,-13.87) -- (3.84,-13.00) -- (2.97,-11.01) -- cycle;
\fill[color=zzttqq,fill=zzttqq,fill opacity=0.1] (3.28,-11.6) -- (4.15,-13.49) -- (6.04,-12.62) -- (5.17,-10.73) -- cycle;
\fill[color=zzttqq,fill=zzttqq,fill opacity=0.1] (5.4,-11.36) -- (6.27,-13.25) -- (8.16,-12.38) -- (7.29,-10.49) -- cycle;
\fill[color=zzttqq,fill=zzttqq,fill opacity=0.1] (0.09,-14.7) -- (0.96,-16.6) -- (2.85,-15.73) -- (1.98,-13.83) -- cycle;
\fill[color=zzttqq,fill=zzttqq,fill opacity=0.1] (5.49,-13.59) -- (6.36,-15.48) -- (8.25,-14.61) -- (7.38,-12.71) -- cycle;
\fill[color=zzttqq,fill=zzttqq,fill opacity=0.1] (4.15,-13.49) -- (5.02,-15.38) -- (6.91,-14.51) -- (6.04,-12.62) -- cycle;
\fill[color=zzttqq,fill=zzttqq,fill opacity=0.1] (2.03,-13.9) -- (2.9,-15.79) -- (4.8,-14.92) -- (3.93,-13.03) -- cycle;
\fill[color=zzttqq,fill=zzttqq,fill opacity=0.1] (3.49,-17.86) -- (4.36,-19.76) -- (6.25,-18.88) -- (5.38,-16.99) -- cycle;
\fill[color=zzttqq,fill=zzttqq,fill opacity=0.1] (1.25,-18.06) -- (2.12,-19.95) -- (4.01,-19.08) -- (3.14,-17.19) -- cycle;
\fill[color=zzttqq,fill=zzttqq,fill opacity=0.1] (3.97,-15.26) -- (4.84,-17.16) -- (6.73,-16.29) -- (5.86,-14.39) -- cycle;
\fill[color=zzttqq,fill=zzttqq,fill opacity=0.1] (2.05,-16.18) -- (2.92,-18.08) -- (4.81,-17.21) -- (3.94,-15.31) -- cycle;
\fill[color=zzttqq,fill=zzttqq,fill opacity=0.1] (0.17,-17.02) -- (1.04,-18.92) -- (2.93,-18.05) -- (2.06,-16.15) -- cycle;
\fill[color=zzttqq,fill=zzttqq,fill opacity=0.1] (-2.31,-16.5) -- (-1.44,-18.4) -- (0.45,-17.53) -- (-0.42,-15.63) -- cycle;
\fill[color=zzttqq,fill=zzttqq,fill opacity=0.1] (-1.19,-15.38) -- (-0.32,-17.28) -- (1.57,-16.41) -- (0.7,-14.51) -- cycle;
\draw [color=qqccww] (0.45,3.23)-- (4.68,3.68);
\draw [color=qqccww] (4.68,3.68)-- (7.26,0.42);
\draw [color=qqccww] (7.26,0.42)-- (4.17,-6.15);
\draw [color=qqccww] (4.17,-6.15)-- (-2.37,-3.77);
\draw [color=qqccww] (-2.37,-3.77)-- (1.9,-0.87);
\draw [color=qqccww] (1.9,-0.87)-- (0.45,3.23);
\draw [color=zzttqq] (0.23,2.99)-- (2.06,1.73);
\draw [color=zzttqq] (2.06,1.73)-- (3.32,3.56);
\draw [color=zzttqq] (3.32,3.56)-- (1.49,4.82);
\draw [color=zzttqq] (1.49,4.82)-- (0.23,2.99);
\draw [color=zzttqq] (3.32,3.56)-- (3.95,1.73);
\draw [color=zzttqq] (3.95,1.73)-- (5.78,2.36);
\draw [color=zzttqq] (5.78,2.36)-- (5.15,4.19);
\draw [color=zzttqq] (5.15,4.19)-- (3.32,3.56);
\draw [color=zzttqq] (5.78,2.36)-- (5.26,0.39);
\draw [color=zzttqq] (5.26,0.39)-- (7.23,-0.13);
\draw [color=zzttqq] (7.23,-0.13)-- (7.75,1.84);
\draw [color=zzttqq] (7.75,1.84)-- (5.78,2.36);
\draw [color=zzttqq] (5.16,-0.68)-- (5.81,-2.63);
\draw [color=zzttqq] (5.81,-2.63)-- (7.77,-1.98);
\draw [color=zzttqq] (7.77,-1.98)-- (7.11,-0.02);
\draw [color=zzttqq] (7.11,-0.02)-- (5.16,-0.68);
\draw [color=zzttqq] (3.86,-3.33)-- (4.53,-5.29);
\draw [color=zzttqq] (4.53,-5.29)-- (6.5,-4.62);
\draw [color=zzttqq] (6.5,-4.62)-- (5.82,-2.65);
\draw [color=zzttqq] (5.82,-2.65)-- (3.86,-3.33);
\draw [color=zzttqq] (1.9,-5.41)-- (3.24,-6.95);
\draw [color=zzttqq] (3.24,-6.95)-- (4.79,-5.6);
\draw [color=zzttqq] (4.79,-5.6)-- (3.44,-4.06);
\draw [color=zzttqq] (3.44,-4.06)-- (1.9,-5.41);
\draw [color=zzttqq] (-0.79,-4.34)-- (0.08,-6.23);
\draw [color=zzttqq] (0.08,-6.23)-- (1.97,-5.36);
\draw [color=zzttqq] (1.97,-5.36)-- (1.1,-3.47);
\draw [color=zzttqq] (1.1,-3.47)-- (-0.79,-4.34);
\draw [color=zzttqq] (-2.31,-2.26)-- (-2.38,-4.31);
\draw [color=zzttqq] (-2.38,-4.31)-- (-0.34,-4.38);
\draw [color=zzttqq] (-0.34,-4.38)-- (-0.27,-2.34);
\draw [color=zzttqq] (-0.27,-2.34)-- (-2.31,-2.26);
\draw [color=zzttqq] (-0.27,-2.34)-- (1.88,-2.33);
\draw [color=zzttqq] (1.88,-2.33)-- (1.88,-0.18);
\draw [color=zzttqq] (1.88,-0.18)-- (-0.27,-0.19);
\draw [color=zzttqq] (-0.27,-0.19)-- (-0.27,-2.34);
\draw [color=zzttqq] (1.65,-0.14)-- (2.53,1.72);
\draw [color=zzttqq] (2.53,1.72)-- (0.67,2.61);
\draw [color=zzttqq] (0.67,2.61)-- (-0.22,0.74);
\draw [color=zzttqq] (-0.22,0.74)-- (1.65,-0.14);
\draw [color=qqccww] (-11.1,3.04)-- (-6.87,3.49);
\draw [color=qqccww] (-6.87,3.49)-- (-4.29,0.23);
\draw [color=qqccww] (-4.29,0.23)-- (-7.38,-6.34);
\draw [color=qqccww] (-7.38,-6.34)-- (-13.92,-3.95);
\draw [color=qqccww] (-13.92,-3.95)-- (-9.65,-1.06);
\draw [color=qqccww] (-9.65,-1.06)-- (-11.1,3.04);
\draw(-6.98,3.45) circle (1.38cm);
\draw(-5.18,1.33) circle (1.38cm);
\draw(-4.88,-0.98) circle (1.38cm);
\draw(-6.04,-3.48) circle (1.38cm);
\draw(-7.79,-5.44) circle (1.38cm);
\draw(-10.49,-5.21) circle (1.38cm);
\draw(-12.6,-3.65) circle (1.38cm);
\draw(-10.66,-1.73) circle (1.38cm);
\draw(-10.06,0.69) circle (1.38cm);
\draw(-9.67,3) circle (1.38cm);
\draw [color=qqccww] (-11.13,-9.59)-- (-6.91,-9.15);
\draw [color=qqccww] (-6.91,-9.15)-- (-4.33,-12.4);
\draw [color=qqccww] (-4.33,-12.4)-- (-7.42,-18.98);
\draw [color=qqccww] (-7.42,-18.98)-- (-13.96,-16.59);
\draw [color=qqccww] (-13.96,-16.59)-- (-9.69,-13.69);
\draw [color=qqccww] (-9.69,-13.69)-- (-11.13,-9.59);
\draw [color=qqccww] (0.59,-9.8)-- (4.81,-9.36);
\draw [color=qqccww] (4.81,-9.36)-- (7.39,-12.61);
\draw [color=qqccww] (7.39,-12.61)-- (4.3,-19.19);
\draw [color=qqccww] (4.3,-19.19)-- (-2.24,-16.8);
\draw [color=qqccww] (-2.24,-16.8)-- (2.03,-13.9);
\draw [color=qqccww] (2.03,-13.9)-- (0.59,-9.8);
\draw(-9.7,-9.28) circle (1.38cm);
\draw(-6.94,-9.46) circle (1.38cm);
\draw(-10.99,-11.17) circle (1.38cm);
\draw(-8.27,-10.71) circle (1.38cm);
\draw(-5.86,-11.76) circle (1.38cm);
\draw(-9.28,-12.07) circle (1.38cm);
\draw(-7.11,-13.12) circle (1.37cm);
\draw(-5.27,-13.31) circle (1.38cm);
\draw(-9.04,-14.48) circle (1.38cm);
\draw(-6.68,-15.37) circle (1.38cm);
\draw(-7.49,-17.65) circle (1.39cm);
\draw(-8.98,-16.66) circle (1.38cm);
\draw(-10.14,-18.06) circle (1.49cm);
\draw(-11.15,-15.29) circle (1.38cm);
\draw(-12.64,-17.04) circle (1.38cm);
\draw [color=zzttqq] (0.21,-9.99)-- (1.08,-11.88);
\draw [color=zzttqq] (1.08,-11.88)-- (2.97,-11.01);
\draw [color=zzttqq] (2.97,-11.01)-- (2.1,-9.12);
\draw [color=zzttqq] (2.1,-9.12)-- (0.21,-9.99);
\draw [color=zzttqq] (2.41,-9.71)-- (3.28,-11.6);
\draw [color=zzttqq] (3.28,-11.6)-- (5.17,-10.73);
\draw [color=zzttqq] (5.17,-10.73)-- (4.3,-8.84);
\draw [color=zzttqq] (4.3,-8.84)-- (2.41,-9.71);
\draw [color=zzttqq] (4.53,-9.47)-- (5.4,-11.36);
\draw [color=zzttqq] (5.4,-11.36)-- (7.29,-10.49);
\draw [color=zzttqq] (7.29,-10.49)-- (6.42,-8.6);
\draw [color=zzttqq] (6.42,-8.6)-- (4.53,-9.47);
\draw [color=zzttqq] (1.08,-11.88)-- (1.95,-13.87);
\draw [color=zzttqq] (1.95,-13.87)-- (3.84,-13.00);
\draw [color=zzttqq] (3.84,-13.00)-- (2.97,-11.01);
\draw [color=zzttqq] (2.97,-11.01)-- (1.08,-11.88);
\draw [color=zzttqq] (3.28,-11.6)-- (4.15,-13.49);
\draw [color=zzttqq] (4.15,-13.49)-- (6.04,-12.62);
\draw [color=zzttqq] (6.04,-12.62)-- (5.17,-10.73);
\draw [color=zzttqq] (5.17,-10.73)-- (3.28,-11.6);
\draw [color=zzttqq] (5.4,-11.36)-- (6.27,-13.25);
\draw [color=zzttqq] (6.27,-13.25)-- (8.16,-12.38);
\draw [color=zzttqq] (8.16,-12.38)-- (7.29,-10.49);
\draw [color=zzttqq] (7.29,-10.49)-- (5.4,-11.36);
\draw [color=zzttqq] (0.09,-14.7)-- (0.96,-16.6);
\draw [color=zzttqq] (0.96,-16.6)-- (2.85,-15.73);
\draw [color=zzttqq] (2.85,-15.73)-- (1.98,-13.83);
\draw [color=zzttqq] (1.98,-13.83)-- (0.09,-14.7);
\draw [color=zzttqq] (5.49,-13.59)-- (6.36,-15.48);
\draw [color=zzttqq] (6.36,-15.48)-- (8.25,-14.61);
\draw [color=zzttqq] (8.25,-14.61)-- (7.38,-12.71);
\draw [color=zzttqq] (7.38,-12.71)-- (5.49,-13.59);
\draw [color=zzttqq] (4.15,-13.49)-- (5.02,-15.38);
\draw [color=zzttqq] (5.02,-15.38)-- (6.91,-14.51);
\draw [color=zzttqq] (6.91,-14.51)-- (6.04,-12.62);
\draw [color=zzttqq] (6.04,-12.62)-- (4.15,-13.49);
\draw [color=zzttqq] (2.03,-13.9)-- (2.9,-15.79);
\draw [color=zzttqq] (2.9,-15.79)-- (4.8,-14.92);
\draw [color=zzttqq] (4.8,-14.92)-- (3.93,-13.03);
\draw [color=zzttqq] (3.93,-13.03)-- (2.03,-13.9);
\draw [color=zzttqq] (3.49,-17.86)-- (4.36,-19.76);
\draw [color=zzttqq] (4.36,-19.76)-- (6.25,-18.88);
\draw [color=zzttqq] (6.25,-18.88)-- (5.38,-16.99);
\draw [color=zzttqq] (5.38,-16.99)-- (3.49,-17.86);
\draw [color=zzttqq] (1.25,-18.06)-- (2.12,-19.95);
\draw [color=zzttqq] (2.12,-19.95)-- (4.01,-19.08);
\draw [color=zzttqq] (4.01,-19.08)-- (3.14,-17.19);
\draw [color=zzttqq] (3.14,-17.19)-- (1.25,-18.06);
\draw [color=zzttqq] (3.97,-15.26)-- (4.84,-17.16);
\draw [color=zzttqq] (4.84,-17.16)-- (6.73,-16.29);
\draw [color=zzttqq] (6.73,-16.29)-- (5.86,-14.39);
\draw [color=zzttqq] (5.86,-14.39)-- (3.97,-15.26);
\draw [color=zzttqq] (2.05,-16.18)-- (2.92,-18.08);
\draw [color=zzttqq] (2.92,-18.08)-- (4.81,-17.21);
\draw [color=zzttqq] (4.81,-17.21)-- (3.94,-15.31);
\draw [color=zzttqq] (3.94,-15.31)-- (2.05,-16.18);
\draw [color=zzttqq] (0.17,-17.02)-- (1.04,-18.92);
\draw [color=zzttqq] (1.04,-18.92)-- (2.93,-18.05);
\draw [color=zzttqq] (2.93,-18.05)-- (2.06,-16.15);
\draw [color=zzttqq] (2.06,-16.15)-- (0.17,-17.02);
\draw [color=zzttqq] (-2.31,-16.5)-- (-1.44,-18.4);
\draw [color=zzttqq] (-1.44,-18.4)-- (0.45,-17.53);
\draw [color=zzttqq] (0.45,-17.53)-- (-0.42,-15.63);
\draw [color=zzttqq] (-0.42,-15.63)-- (-2.31,-16.5);
\draw [color=zzttqq] (-1.19,-15.38)-- (-0.32,-17.28);
\draw [color=zzttqq] (-0.32,-17.28)-- (1.57,-16.41);
\draw [color=zzttqq] (1.57,-16.41)-- (0.7,-14.51);
\draw [color=zzttqq] (0.7,-14.51)-- (-1.19,-15.38);
\begin{scriptsize}
\end{scriptsize}
\end{tikzpicture}
}

\caption{Coverage footprint of the region boundary or interior with circle or rectangle coverage footprints.}
\label{fig:intro}
\end{figure}

Unfortunately, the memory resource and picture transmission bandwidth on the drone are usually limited, 
so let us assume the drone can only take or transmit back at most $k$ pictures
in one flight, where $k$ is a parameter induced by the drone hardware or the specific region's transmission bandwidth restriction. 
We want to know what the largest zoom factor $\lambda$ is that we can set 
such that a drone can still see all parts of the region with $k$ pictures taken downwards from a gimbal camera. 
Figure~\ref{fig:intro} illustrates a coverage result 
for covering the boundary or interior of a region with circles or squares, 
where circles represent footprints of visible regions using downwards-facing fisheye cameras or radars
and squares represent the specialization of rectangular images taken by regular cameras. 

The theoretical aspect of the above aerial photography problems 
strongly relates to the area of computational geometry, where coverage of 
point sets (a.k.a.\ geometric point clouds) is studied extensively.
For example, the $k$-center problem asks: given a set of points and a number $k$,
what is the minimum radius of $k$ circles centered at a subset of these points 
to cover all the points \cite{toth2017handbook}. 
Similar problems include the $k$-median problem, which asks to minimize the sum of the distances from each point to the closest 
chosen point among the $k$ chosen points,
and the $k$-means problem, which minimizes the sum of squared distances from each point to the closest chosen point. 
These problems, especially $k$-means, can go to high dimensions in machine learning.
However, in real-world aerial photography coverage problems, 
the point sets are associated with specific physical meanings,
and are usually inside the $3$-dimensional Euclidean space in the simple case,
or go to $5$ dimensions to include the normal directions. 

In reality, most of the interesting regions to cover in aerial coverage problems have 
some degree of continuity, for example, a complete structure, a continuous boundary, or a fence.
Continuous geometric shapes like 2D polygons or 3D meshes create more challenges for the usually studied
point-set coverage problems while making the problems more realistic.
In robotics, continuity is usually implied, as a robot handling coverage or 
inspection tasks moves in a continuous manner, and it makes sense to assume the region to cover is
a continuous region instead of a scattered point set.
Moreover, restricting the region to cover to a simple polygon has practical implications for drones.
For multiple regions to cover, the complement of these regions usually represents inaccessible regions to drones or for drones to pass through,
such as high mountains, large ponds, and hazardous areas where landing is impossible in case of emergency.
Hence, drones usually do not cross regions to execute tasks, and it is natural to study the picture-taking effort for a single region.

Early studies on covering and packing \cite{Hachbaum1985Approximation} provide strong
hardness results and approximation schemes for covering points.
Later work \cite{feder1988Optimal} further showed a $1.88$-inapprox.\ gap for point-set coverage using the $L_2$ metric,
and a $2$-inapprox.\ gap for coverage using the $L_1$ metric. 
For continuous regions and boundaries of a simple polygon, recent work \cite{Feng2020Optimally}
has shown a $1.152$-inapprox.\ gap
with the minimum circle radius using $k$ circles.

Coverage in 2D is a simplification of the real-world coverage problem when the region scale is large. 
Meanwhile, covering 3D surfaces has been a popular aerial robotics application studied extensively.
For example, \cite{Roberts2017Submodular} plans an optimal path
for capturing the surface of a structure for 3D reconstruction with the help of
submodular functions and integer programming.
Similarly, \cite{Feng2021Sensor} developed a sampling followed by integer programming
method for computing sensing points to place sensors for complete coverage of a 3D structure,
as well as a $(2+\epsilon)$-optimal algorithm to solve the problem in polynomial time.

In this work, the complexity analysis for covering a simple planar polygon using squares is the main focus,
which shows an inapprox.\ gap of $1.165$ for
finding the minimum side length for $k$ squares to cover the simple 2D region (a simple polygon) and an
inapprox.\ gap of $1.25$ when the square centers are restricted to be inside the region,
and develops a $2.828$-optimal algorithm to solve the square coverage problem.
Since most regular cameras used by gimbals take rectangular images, 
the results in this work bring the computational complexity results in \cite{Feng2020Optimally}
closer to reality for aerial photography problems. 
They can potentially serve as a reference for setting camera zoom factors or designing gimbal cameras
for aerial surveying or mapping. 

Additionally, as the drone's position may be restricted to be only on top of the region to 
take pictures of the region itself, this work also considers the variant of the problem 
when the positions of the square or circle centers must be inside the simply connected region 
or on top of the region's perimeter, where the inapprox.\ gap is the same as in the non-restricted case. 
This variant gives the work in \cite{Feng2019Efficient} an 
aerial photography interpretation.

\section{Problem Formulations}
This section provides the mathematical formulations of the problems studied in this work.

\begin{problem}[Circle Coverage]\label{p:cir_cov}
Given a region, represented by a simple polygon (a planar polygonal chain without self-intersections or holes) $\mathcal{P}$,
and a number $k\geq 1$,
what is the minimum length $\ell$ such that we can place $k$ circles each
with radius $\ell$ so that $\mathcal{P}$ is contained in these circles?
\end{problem}

\begin{problem}[Square Coverage]\label{p:sq_cov}
Given a region (a simple polygon $\mathcal{P}$)
and a number $k\geq 1$, what is the minimum length $\ell$ such that we can place $k$ squares each
with side length $\ell$ so that $\mathcal{P}$ is contained in these squares?
\end{problem}

\begin{remark}
If we only care about covering the boundary of a simple polygon, similar to 
the definition of the barrier coverage problem \cite{Gage1992Command},
this problem can be reduced to the two problems aforementioned because we can solidify the boundary 
by expanding the edges of a polygon to sticks with a small width $\delta$, 
and leave a narrow opening of width $\delta$ for hole elimination. 
When $\delta\rightarrow 0$, the results of coverage problems on the solidified boundary skeleton and the boundary skeleton converge to the same value. 
In this way, covering the expanded polygon from the original perimeter is the same as covering the boundary of the original polygon.
Conversely, a simple polygon can be approximately represented by a polygon whose boundary almost fills in the
interior of the polygon.
Hence, the computational complexity of covering the interior and the boundary of a simple region is of the same level.
\end{remark}

In many real scenarios, the sensors' or the robots' locations (i.e., the centers of the circles or rectangles) 
must be inside the region itself or on the perimeter; 
for example, the watchtowers should be built on or inside the defensive wall itself and not outside; 
unmanned aerial vehicles for security should not go outside or inside the security zone lines.
Because a square is a special case of a rectangle, the complexity analysis in this paper for using rectangles to cover a region
is restricted to using squares to cover a region for ease of the complexity analysis 
(note that using rectangles for the coverage problem can only make the problem harder).
When the constraint that the circle or square centers must be inside the polygon needs to be satisfied, 
the constrained versions of Problems~\ref{p:cir_cov} and~\ref{p:sq_cov} are defined as follows:

\begin{problem}[Constrained Circle or Square Coverage]
\label{p:con_sq_cir_cov}
Same as Problem~\ref{p:cir_cov} and Problem~\ref{p:sq_cov}, but the circle or square centers must stay inside the polygon $\mathcal{P}$.
\end{problem}

\begin{remark}
Similarly, the case when the feasible region to place circle centers is the interior of the polygon is equivalent to the case 
when the feasible region is on the boundary of a polygon,
because we can easily make a polygon out of the boundary of a simple polygon by solidifying the line segments of the polygon
and making narrow openings for hole elimination.
\end{remark}

\section{Preliminaries}\label{sec:preliminaries}
To establish the computational complexity of Problems~\ref{p:cir_cov} through~\ref{p:con_sq_cir_cov}, 
preliminary results and gadget construction steps are introduced in this section.

\begin{definition} [Planar graph]
A graph $G(V, E)$ is planar when we can embed the vertices and edges onto a plane without edges crossing each other.
\end{definition}

\begin{definition} [Cubic graph]
A graph $G(V, E)$ is cubic when each vertex in $G$ has degree $3$.
\end{definition}
    
\begin{problem} [Vertex Cover on Planar Cubic Graph]\label{p:vc_cubic}
Given a planar cubic graph $G(V, E)$ and a number $n$, does there exist $V'\subset V$ such that
$|V'| = n$ and for each $e=(u,v)\in E$, either $u\in V'$ or $v \in V'$?
\end{problem}
Problem~\ref{p:vc_cubic} is shown to be computationally intractable in \cite{Mohar2001Face}.

We will then transform the planar cubic graph into a structure to prove the computational complexity.

\subsection{Intermediate structure construction} \label{subsec:gadgets}
Given a planar cubic graph $G(V, E)$, we convert it into a special structure through a sequence of steps.
The first step embeds the graph $G$ into the plane.
Each vertex in $G$ becomes a vertex junction, and each edge becomes a fence-like structure connecting two 
neighboring junctions, which is an odd-length path intersected with a set of perpendicular bars of length $\zeta$
($\zeta$ is a parameter to be determined for proving different results)
at each unit line segment.

Starting from a planar cubic graph $G$, we construct a structure, $T_G$, as follows.
First, similar to \cite{feder1988Optimal}, to embed $G$ into the plane,
an edge $uw \in E(G)$ is converted to an odd length path $uv_1, v_1v_2,
    \ldots, v_{2m}w$ where $m > 3$ is an integer. We note that $m$ is different
in general for different edges of $G$.
Denote such a path as $u\cdots w$; each edge along $u\cdots w$ is straight
and has unit edge length. We also require that each path is nearly straight
locally.
%
For a vertex of $G$ with degree $3$, e.g., a vertex $u \in V(G)$
neighboring $w, x, y \in V(G)$, we choose proper configurations and lengths for
paths, $u\cdots w$, $u\cdots x$, and $u\cdots y$ such that
these paths meet at $u$ forming pairwise angles of $120^\circ$. We denote the
resulting graph as $G'$, which becomes the {\em backbone} of the $T_G$.

From here, the second modification is made which completes the
construction of $T_G$. In each previously constructed
path $u\cdots w = uv_1\ldots v_{2m}w$, for each $v_iv_{i+1}$, $1 \le i \le 2m-1$, 
we add a line segment of length $\zeta$ that is
perpendicular to $v_iv_{i+1}$ such that $v_iv_{i+1}$ and the line
segment divide each other in the middle. 
$G'$ and the bars form the intermediate structure $T_G$.

\section{Hardness of Approximate Circle Coverage}\label{sec:complexity-circle-coverage}
The theorem in this section is proved completely in \cite{Feng2020Optimally}, but to ease 
the following derivations, we give a simplified version of the proof.

\begin{figure}[h]
    \centering
    \definecolor{xdxdff}{rgb}{0.49,0.49,1}
\definecolor{qqqqff}{rgb}{0,0,1}
\resizebox{\linewidth}{!}{
\begin{tikzpicture}[line cap=round,line join=round,>=triangle 45,x=1.0cm,y=1.0cm]
\clip(-10.59,-19.25) rectangle (20.33,6);
\draw (0,1.73)-- (0,-1.73);
\draw [domain=-6.97:15] plot(\x,{(-0-0*\x)/22});
\draw (2,1.73)-- (2,-1.73);
\draw (4,1.73)-- (4,-1.73);
\draw (6,1.73)-- (6,-1.73);
\draw (1.03,0) circle (1.98cm);
\draw (5.03,0) circle (1.99cm);
\draw (0,-7.85)-- (0,-11.31);
\draw [domain=-6.97:15] plot(\x,{(-210.9--0.05*\x)/21.91});
\draw (2,-7.85)-- (2,-11.31);
\draw (4,-7.85)-- (4,-11.31);
\draw (6,-7.85)-- (6,-11.31);
\draw (7.16,-9.59) circle (2.09cm);
\draw (8.03,1.77)-- (8.03,-1.69);
\draw (10.03,1.77)-- (10.03,-1.69);
\draw (12.03,1.77)-- (12.03,-1.69);
\draw(9.02,0.04) circle (2cm);
\draw(13.03,0.04) circle (2cm);
\draw (-3.96,1.72)-- (-3.96,-1.75);
\draw (-1.96,1.72)-- (-1.96,-1.74);
\draw(-2.96,-0.01) circle (2cm);
\draw (-3.96,-7.88)-- (-3.96,-11.34);
\draw(-4.87,-9.65) circle (1.99cm);
\draw (10.19,-7.85)-- (10.19,-11.31);
\draw (12.19,-7.85)-- (12.19,-11.31);
\draw(11.28,-9.62) circle (1.99cm);
\draw (-1.9,-7.82)-- (-1.9,-11.28);
\draw (8.33,-7.82)-- (8.33,-11.28);
\draw(3.04,-9.59) circle (2.03cm);
\draw(-0.96,-9.56) circle (1.98cm);
\begin{scriptsize}
\fill [color=qqqqff] (15,0) circle (1.5pt);
\draw[color=qqqqff] (15.5,0.43) node {$B_1$};
\fill [color=qqqqff] (0,1.73) circle (1.5pt);
\draw[color=qqqqff] (0.48,2.16) node {$E_2$};
\fill [color=qqqqff] (0,-1.73) circle (1.5pt);
\draw[color=qqqqff] (0.48,-1.31) node {$G_2$};
\fill [color=qqqqff] (2,1.73) circle (1.5pt);
\draw[color=qqqqff] (2.47,2.16) node {$E_1$};
\fill [color=qqqqff] (2,-1.73) circle (1.5pt);
\draw[color=qqqqff] (4.5,-1.31) node {$G_3$};
\fill [color=qqqqff] (6,1.73) circle (1.5pt);
\draw[color=qqqqff] (6.5,2.16) node {$E_4$};
\fill [color=qqqqff] (6,-1.73) circle (1.5pt);
\draw[color=qqqqff] (6.5,-1.31) node {$G_4$};
\fill [color=xdxdff] (1.03,0) circle (1.5pt);
\draw[color=xdxdff] (1.3,0.43) node {$A$};
\fill [color=xdxdff] (5.03,0) circle (1.5pt);
\draw[color=xdxdff] (5.29,0.43) node {$B$};
\fill [color=qqqqff] (-6.91,-9.64) circle (1.5pt);
\draw[color=qqqqff] (-6.43,-9.23) node {$C_1$};
\fill [color=qqqqff] (15,-9.59) circle (1.5pt);
\draw[color=qqqqff] (15.5,-9.16) node {$B_3$};
\fill [color=qqqqff] (0,-7.85) circle (1.5pt);
\draw[color=qqqqff] (0.48,-7.43) node {$E_6$};
\fill [color=qqqqff] (0,-11.31) circle (1.5pt);
\draw[color=qqqqff] (0.48,-10.9) node {$G_6$};
\fill [color=qqqqff] (2,-7.85) circle (1.5pt);
\draw[color=qqqqff] (2.47,-7.43) node {$E_5$};
\fill [color=qqqqff] (2,-11.31) circle (1.5pt);
\draw[color=qqqqff] (2.47,-10.9) node {$G_5$};
\fill [color=qqqqff] (6,-7.85) circle (1.5pt);
\draw[color=qqqqff] (6.5,-7.43) node {$E_8$};
\fill [color=qqqqff] (6,-11.31) circle (1.5pt);
\draw[color=qqqqff] (6.5,-10.9) node {$G_8$};
\fill [color=xdxdff] (7.16,-9.59) circle (1.5pt);
\draw[color=xdxdff] (7.65,-9.16) node {$B_4$};
\fill [color=qqqqff] (8.03,1.77) circle (1.5pt);
\draw[color=qqqqff] (8.92,2.2) node {$E_{10}$};
\fill [color=qqqqff] (8.03,-1.69) circle (1.5pt);
\draw[color=qqqqff] (8.53,-1.27) node {$G_9$};
\fill [color=qqqqff] (10.03,1.77) circle (1.5pt);
\draw[color=qqqqff] (10.92,2.2) node {$E_{11}$};
\fill [color=qqqqff] (10.03,-1.69) circle (1.5pt);
\draw[color=qqqqff] (10.92,-1.27) node {$G_{10}$};
\fill [color=qqqqff] (12.03,1.77) circle (1.5pt);
\draw[color=qqqqff] (12.92,2.2) node {$E_{12}$};
\fill [color=qqqqff] (12.03,-1.69) circle (1.5pt);
\draw[color=qqqqff] (12.92,-1.27) node {$G_{11}$};
\fill [color=xdxdff] (9.02,0.04) circle (1.5pt);
\draw[color=xdxdff] (9.51,0.46) node {$A_4$};
\fill [color=xdxdff] (13.03,0.04) circle (1.5pt);
\draw[color=xdxdff] (13.54,0.46) node {$B_7$};
\fill [color=qqqqff] (-3.96,1.72) circle (1.5pt);
\draw[color=qqqqff] (-3.09,2.13) node {$E_{16}$};
\fill [color=qqqqff] (-3.96,-1.75) circle (1.5pt);
\draw[color=qqqqff] (-3.09,-1.31) node {$G_{15}$};
\fill [color=qqqqff] (-1.96,1.72) circle (1.5pt);
\draw[color=qqqqff] (-1.09,2.13) node {$E_{17}$};
\fill [color=qqqqff] (-1.96,-1.74) circle (1.5pt);
\draw[color=qqqqff] (-1.09,-1.31) node {$G_{16}$};
\fill [color=xdxdff] (-6.97,-0.01) circle (1.5pt);
\draw[color=xdxdff] (-6.46,0.43) node {$A_5$};
\fill [color=xdxdff] (-2.96,-0.01) circle (1.5pt);
\draw[color=xdxdff] (-2.47,0.43) node {$B_9$};
\fill [color=qqqqff] (-3.96,-7.88) circle (1.5pt);
\draw[color=qqqqff] (-3.09,-7.46) node {$E_{18}$};
\fill [color=qqqqff] (-3.96,-11.34) circle (1.5pt);
\draw[color=qqqqff] (-3.09,-10.93) node {$G_{17}$};
\fill [color=xdxdff] (-4.87,-9.65) circle (1.5pt);
\draw[color=xdxdff] (-3.97,-9.23) node {$B_{11}$};
\fill [color=qqqqff] (10.19,-7.85) circle (1.5pt);
\draw[color=qqqqff] (11.08,-7.43) node {$E_{20}$};
\fill [color=qqqqff] (10.19,-11.31) circle (1.5pt);
\draw[color=qqqqff] (11.08,-10.9) node {$G_{18}$};
\fill [color=qqqqff] (12.19,-7.85) circle (1.5pt);
\draw[color=qqqqff] (13.08,-7.43) node {$E_{21}$};
\fill [color=qqqqff] (12.19,-11.31) circle (1.5pt);
\draw[color=qqqqff] (13.08,-10.9) node {$G_{19}$};
\fill [color=xdxdff] (11.28,-9.62) circle (1.5pt);
\draw[color=xdxdff] (12.16,-9.2) node {$B_{13}$};
\fill [color=qqqqff] (-1.9,-7.82) circle (1.5pt);
\draw[color=qqqqff] (-1.03,-7.4) node {$E_{23}$};
\fill [color=qqqqff] (-1.9,-11.28) circle (1.5pt);
\draw[color=qqqqff] (-1.03,-10.87) node {$G_{20}$};
\fill [color=qqqqff] (8.33,-7.82) circle (1.5pt);
\draw[color=qqqqff] (9.22,-7.4) node {$E_{14}$};
\fill [color=qqqqff] (8.33,-11.28) circle (1.5pt);
\draw[color=qqqqff] (9.22,-10.87) node {$G_{12}$};
\fill [color=xdxdff] (3.04,-9.59) circle (1.5pt);
\draw[color=xdxdff] (3.52,-9.16) node {$B_2$};
\fill [color=xdxdff] (-0.96,-9.56) circle (1.5pt);
\draw[color=xdxdff] (-0.08,-9.13) node {$B_{16}$};

\end{scriptsize}
\end{tikzpicture}}
    \caption[Structure within the odd length path and attached perpendicular ``bars'']
    {Structure within the odd length path and attached
        perpendicular ``bars'' with length $\zeta = \sqrt{3}$. Regarding the
        representation of such non-integral coordinates in the problem
        input,
        we may scale the coordinates to some certain extent and
        round them to integers so that the relative distance between
        each other is precise enough
        for the proof.
    }
    \vspace*{-1mm}
    \label{fig:osg-path-bar}
\end{figure}
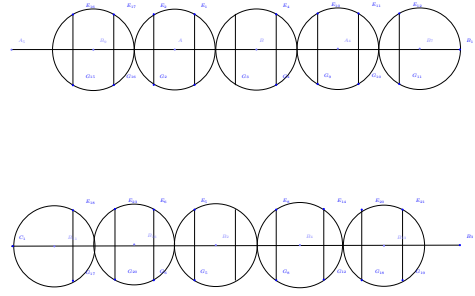

\begin{figure}[h]
    \centering\definecolor{qqqqff}{rgb}{0,0,1}
\resizebox{\linewidth}{!}{
\begin{tikzpicture}[line cap=round,line join=round,x=1.0cm,y=1.0cm]
\clip(-11.1,-7.88) rectangle (11.28,7.83);
\draw [domain=-8.096441996930727:0.0] plot(\x,{(-0--17.32*\x)/-10});
\draw [domain=-8.096441996930727:0.0] plot(\x,{(-0-17.32*\x)/-10});
\draw [domain=0.0:8.282611203888568] plot(\x,{(-0-0*\x)/20});
\draw [dash pattern=on 5pt off 5pt] (107.4,-8.88) -- (107.4,8.83);
\draw (3,1.73)-- (3,-1.73);
\draw(2,-0.04) circle (2cm);
\draw (5,1.73)-- (5,-1.73);
\draw(-1.04,1.71) circle (2cm);
\draw (0,3.46)-- (-3,1.73);
\draw (-1,5.2)-- (-4,3.46);
\draw (-3,-1.73)-- (0,-3.46);
\draw (-4,-3.46)-- (-1,-5.2);
\draw(-0.96,-1.74) circle (2cm);
\begin{scriptsize}
\fill [color=qqqqff] (0,0) circle (1.5pt);
\draw[color=qqqqff] (0.15,0.24) node {$C$};
\fill [color=qqqqff] (-10,17.32) circle (1.5pt);
\draw[color=qqqqff] (-11.02,8.96) node {$D$};
\fill [color=qqqqff] (-10,-17.32) circle (1.5pt);
\draw[color=qqqqff] (-11.02,8.96) node {$A$};
\fill [color=qqqqff] (20,0) circle (1.5pt);
\draw[color=qqqqff] (-11.02,8.96) node {$B$};
\fill [color=qqqqff] (107.4,-62.09) circle (1.5pt);
\draw[color=qqqqff] (-11.02,8.96) node {$E$};
\fill [color=qqqqff] (3,1.73) circle (1.5pt);
\draw[color=qqqqff] (3.16,1.98) node {$G$};
\fill [color=qqqqff] (3,-1.73) circle (1.5pt);
\draw[color=qqqqff] (3.16,-1.47) node {$H$};
\fill [color=qqqqff] (5,1.73) circle (1.5pt);
\draw[color=qqqqff] (5.28,1.98) node {$G_1$};
\fill [color=qqqqff] (5,-1.73) circle (1.5pt);
\draw[color=qqqqff] (5.28,-1.47) node {$H_1$};
\fill [color=qqqqff] (0,3.46) circle (1.5pt);
\draw[color=qqqqff] (0.19,3.72) node {$G'$};
\fill [color=qqqqff] (-3,1.73) circle (1.5pt);
\draw[color=qqqqff] (-2.8,1.98) node {$H'$};
\fill [color=qqqqff] (-1,5.2) circle (1.5pt);
\draw[color=qqqqff] (-0.91,5.45) node {$I$};
\fill [color=qqqqff] (-4,3.46) circle (1.5pt);
\draw[color=qqqqff] (-3.9,3.72) node {$J$};
\fill [color=qqqqff] (-3,-1.73) circle (1.5pt);
\draw[color=qqqqff] (-2.66,-1.47) node {$G'_1$};
\fill [color=qqqqff] (0,-3.46) circle (1.5pt);
\draw[color=qqqqff] (0.33,-3.21) node {$H'_1$};
\fill [color=qqqqff] (-4,-3.46) circle (1.5pt);
\draw[color=qqqqff] (-3.86,-3.21) node {$K$};
\fill [color=qqqqff] (-1,-5.2) circle (1.5pt);
\draw[color=qqqqff] (-0.87,-4.95) node {$L$};
\end{scriptsize}
\end{tikzpicture}}
\caption{Circle coverage pattern at the junction crossing with unit radius.}
\label{fig:osg-crossing-unit-cir}
\end{figure}
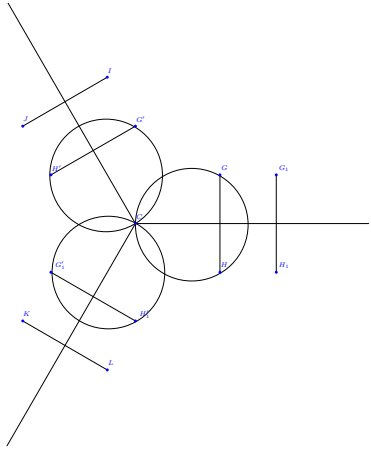

\begin{figure}[h]
\definecolor{qqqqff}{rgb}{0,0,1}
\resizebox{\linewidth}{!}{
\begin{tikzpicture}[line cap=round,line join=round,x=1.0cm,y=1.0cm]
\clip(-8.39,-7.4) rectangle (8.27,6.61);
\draw [domain=-8.385323169122886:0.0] plot(\x,{(-0--17.32*\x)/-10});
\draw [domain=-8.385323169122886:0.0] plot(\x,{(-0-17.32*\x)/-10});
\draw [domain=0.0:8.269718461816346] plot(\x,{(-0-0*\x)/20});
\draw [dash pattern=on 4pt off 4pt] (107.4,-7.4) -- (107.4,6.61);
\draw (3,1.73)-- (3,-1.73);
\draw (5,1.73)-- (5,-1.73);
\draw (0,3.46)-- (-3,1.73);
\draw (-1,5.2)-- (-4,3.46);
\draw (-3,-1.73)-- (0,-3.46);
\draw (-4,-3.46)-- (-1,-5.2);
\draw(-0.71,1.24) circle (2.31cm);
\draw(1.46,-0.02) circle (2.31cm);
\draw(-1.73,-3.07) circle (2.31cm);
\begin{scriptsize}
\fill [color=qqqqff] (0,0) circle (1.5pt);
\draw[color=qqqqff] (0.12,0.2) node {$C$};
\fill [color=qqqqff] (-10,17.32) circle (1.5pt);
\draw[color=qqqqff] (-8.32,6.72) node {$D$};
\fill [color=qqqqff] (-10,-17.32) circle (1.5pt);
\draw[color=qqqqff] (-8.32,6.72) node {$A$};
\fill [color=qqqqff] (20,0) circle (1.5pt);
\draw[color=qqqqff] (-8.32,6.72) node {$B$};
\fill [color=qqqqff] (107.4,-62.09) circle (1.5pt);
\draw[color=qqqqff] (-8.32,6.72) node {$E$};
\fill [color=qqqqff] (3,1.73) circle (1.5pt);
\draw[color=qqqqff] (3.12,1.94) node {$G$};
\fill [color=qqqqff] (3,-1.73) circle (1.5pt);
\draw[color=qqqqff] (3.12,-1.53) node {$H$};
\fill [color=qqqqff] (5,1.73) circle (1.5pt);
\draw[color=qqqqff] (5.23,1.94) node {$G_1$};
\fill [color=qqqqff] (5,-1.73) circle (1.5pt);
\draw[color=qqqqff] (5.23,-1.53) node {$H_1$};
\fill [color=qqqqff] (0,3.46) circle (1.5pt);
\draw[color=qqqqff] (0.15,3.67) node {$G'$};
\fill [color=qqqqff] (-3,1.73) circle (1.5pt);
\draw[color=qqqqff] (-2.83,1.94) node {$H'$};
\fill [color=qqqqff] (-1,5.2) circle (1.5pt);
\draw[color=qqqqff] (-0.92,5.4) node {$I$};
\fill [color=qqqqff] (-4,3.46) circle (1.5pt);
\draw[color=qqqqff] (-3.92,3.67) node {$J$};
\fill [color=qqqqff] (-2.01,3.47) circle (1.5pt);
\draw[color=qqqqff] (-1.86,3.67) node {$F'$};
\fill [color=qqqqff] (-3,-1.73) circle (1.5pt);
\draw[color=qqqqff] (-2.74,-1.53) node {$G'_1$};
\fill [color=qqqqff] (0,-3.46) circle (1.5pt);
\draw[color=qqqqff] (0.26,-3.26) node {$H'_1$};
\fill [color=qqqqff] (-4,-3.46) circle (1.5pt);
\draw[color=qqqqff] (-3.89,-3.26) node {$K$};
\fill [color=qqqqff] (-1,-5.2) circle (1.5pt);
\draw[color=qqqqff] (-0.88,-4.99) node {$L$};
\fill [color=qqqqff] (-2,-3.47) circle (1.5pt);
\draw[color=qqqqff] (-1.75,-3.28) node {$F'_1$};

\fill [color=qqqqff] (-0.62,-1.10) circle (1.5pt);
\draw[color=qqqqff] (-0.37,-0.9) node {$G$};

\fill [color=qqqqff] (-1.73,-3.07) circle (1.5pt);
\draw[color=qqqqff] (-1.51,-2.87) node {$F_4$};

\fill [color=qqqqff] (-2.5,-4.35) circle (1.5pt);
\draw[color=qqqqff] (-2.3,-4.15) node {$I$};

\fill [color=qqqqff] (1.6,0) circle (1.5pt);
\draw[color=qqqqff] (1.9,0.3) node {$F_5$};

\fill [color=qqqqff] (3.0,0) circle (1.5pt);
\draw[color=qqqqff] (3.3,0.3) node {$I'$};

\end{scriptsize}
\end{tikzpicture}}
\caption{Circle coverage pattern at the extreme circle radius at the junction crossing.}
\label{fig:junction_crossing}
\end{figure}
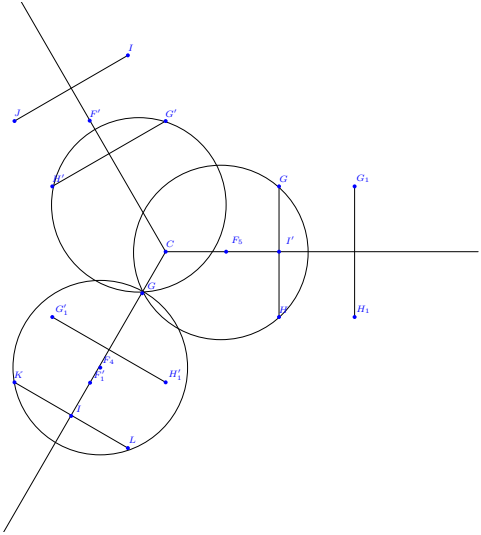

The gadget structure (refer to it as $T$) constructed in the previous section can be easily converted to a simple polygon
by dilating the lines to a thin structure with a small width $\varepsilon$ while leaving a narrow opening of width $\varepsilon$ 
to make it a simple polygon (refer to it as $P_\varepsilon$). Assume the minimum radius of $k$ circles to cover a simple polygon is
$\ell^*$; then the minimum radius $\ell$ for covering $P_\varepsilon$ must satisfy 
$\ell - \varepsilon \leq \ell^* \leq \ell + \varepsilon$. 
When $\varepsilon \rightarrow 0$, $\ell$ and $\ell^*$ converge to the same value.
So, we only need to show the complexity of approximating the placement of $k$ circles of minimum radius to cover the structure $T$.

First, we set the bar length $\zeta=\sqrt{3}$ in the gadget structure.
Clearly, for each edge link $u v_1\dots v_{2i} w$ between junctions $u$ and $w$, to cover the link with the minimum number 
of circles when the circle radius is less than $1.155$ \cite{Feng2020Optimally}, 
one of the two patterns in Figure~\ref{fig:osg-path-bar} should be used. 

Since each junction maps to a vertex in $G$ in the transformation, call the side of the edge with a circle covering only one bar
the ``odd-end'', and the other side of the edge the ``even-end''. 
If there is a vertex cover of graph $G$, then it is possible to let all the chosen vertices be the ``even-end'' or ``odd-end'',
and the rest of the vertices be ``even-end'' in the edge-link coverage pattern. Since every junction-represented vertex 
is either inside a vertex cover or has a neighbor inside the vertex cover, it is always possible to find such an ``odd-end'' and ``even-end''
combination for the junctions so that at least no junction has all three edges as even ends.

On the other hand, for a coverage of $T$ with the minimum number of circles, select those vertices
with at least one even-end. It is clear that these vertices form a vertex cover. 


At the maximum circle radius $\ell$ such that the structure still needs the same number of circles to cover
as when using unit circles, Figure~\ref{fig:junction_crossing} shows this situation when the junction can be covered
by two circles covering one vertical bar each and another circle covering two vertical bars.

Listing all necessary geometric constraints in Figure~\ref{fig:junction_crossing} gives
\begin{align}
    \label{eq:sq_crossing_start}
    \|CG\| + \|F_4G\| + \|F_4 I\| &= 1.75 \\
    \|F_4G\| = \|F_5G\| &= \ell \\
    \|IL\| = \|GI'\| &= \sqrt{3} / 2 \\
    \|F_4 I \|^2 + \|IL\|^2 &= \ell^2 \\
    \|F_5 I'\|^2 + \|GI'\|^2 &= \ell ^2 \\
    \|CG\|^2 + \|CF_5\|^2 + \|CG\|\cdot \|CF_5\| &= \|F_5 G\|^2
    \label{eq:sq_crossing_end}
\end{align}

Solving the system of equations from \eqref{eq:sq_crossing_start} to \eqref{eq:sq_crossing_end} with SimPy \cite{Meurer2017Sympy} gives
$\ell \approx 1.152$,
which is the extreme case when the circle coverage pattern of $T_G$ cannot map to a vertex
cover of the original vertex cover problem.

\begin{theorem}\cite{Feng2020Optimally}
    It is NP-hard to approximate Problem~\ref{p:cir_cov} within a factor of $1.152$.
\end{theorem}

\section{Hardness of Approximate Square Coverage}\label{sec:complexity-square-coverage}
In the structure gadget constructed in the previous section, 
let us shrink the unit-length segment to $\sqrt{2}/2$, 
so the current distance between neighboring bars becomes ${\sqrt{2}} / {2}$.
And the bar length $\zeta$ here is set to be $\zeta = {\sqrt{2}}/{2}$. 

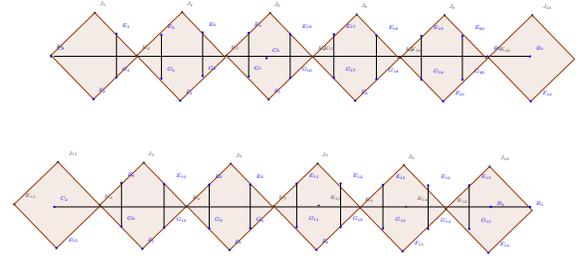
\begin{figure}[ht]
\definecolor{uququq}{rgb}{0.25,0.25,0.25}
\definecolor{zzttqq}{rgb}{0.6,0.2,0}
\definecolor{qqqqff}{rgb}{0,0,1}
\centering
\resizebox{\linewidth}{!}{
\begin{tikzpicture}[line cap=round,line join=round,x=1.0cm,y=1.0cm]
\clip(0.03,-23.18) rectangle (28.94,-10.08);
\fill[color=zzttqq,fill=zzttqq,fill opacity=0.1] (4.74,-15.17) -- (6.75,-13.22) -- (4.8,-11.21) -- (2.79,-13.16) -- cycle;
\fill[color=zzttqq,fill=zzttqq,fill opacity=0.1] (8.71,-15.24) -- (10.78,-13.26) -- (8.8,-11.18) -- (6.72,-13.17) -- cycle;
\fill[color=zzttqq,fill=zzttqq,fill opacity=0.1] (6.98,-22.03) -- (8.99,-20.09) -- (7.04,-18.08) -- (5.03,-20.02) -- cycle;
\fill[color=zzttqq,fill=zzttqq,fill opacity=0.1] (10.97,-22.09) -- (12.98,-20.14) -- (11.03,-18.13) -- (9.02,-20.08) -- cycle;
\fill[color=zzttqq,fill=zzttqq,fill opacity=0.1] (12.76,-15.18) -- (14.77,-13.23) -- (12.83,-11.22) -- (10.82,-13.17) -- cycle;
\fill[color=zzttqq,fill=zzttqq,fill opacity=0.1] (16.73,-15.24) -- (18.74,-13.3) -- (16.8,-11.29) -- (14.79,-13.23) -- cycle;
\fill[color=zzttqq,fill=zzttqq,fill opacity=0.1] (14.95,-22.08) -- (16.96,-20.13) -- (15.01,-18.12) -- (13,-20.07) -- cycle;
\fill[color=zzttqq,fill=zzttqq,fill opacity=0.1] (20.76,-15.27) -- (22.77,-13.33) -- (20.82,-11.32) -- (18.81,-13.26) -- cycle;
\fill[color=zzttqq,fill=zzttqq,fill opacity=0.1] (18.9,-22.15) -- (20.91,-20.21) -- (18.96,-18.2) -- (16.95,-20.14) -- cycle;
\fill[color=zzttqq,fill=zzttqq,fill opacity=0.1] (24.77,-15.27) -- (26.78,-13.33) -- (24.84,-11.32) -- (22.83,-13.26) -- cycle;
\fill[color=zzttqq,fill=zzttqq,fill opacity=0.1] (3.04,-22) -- (5.05,-20.06) -- (3.11,-18.05) -- (1.1,-19.99) -- cycle;
\fill[color=zzttqq,fill=zzttqq,fill opacity=0.1] (22.83,-22.21) -- (24.84,-20.27) -- (22.89,-18.26) -- (20.88,-20.2) -- cycle;
\draw (5.79,-12.17)-- (5.79,-14.17);
\draw (7.85,-12.22)-- (7.85,-14.22);
\draw (9.74,-12.11)-- (9.74,-14.11);
\draw [color=zzttqq] (4.74,-15.17)-- (6.75,-13.22);
\draw [color=zzttqq] (6.75,-13.22)-- (4.8,-11.21);
\draw [color=zzttqq] (4.8,-11.21)-- (2.79,-13.16);
\draw [color=zzttqq] (2.79,-13.16)-- (4.74,-15.17);
\draw [domain=2.94:24.74] plot(\x,{(-264.11-0*\x)/20});
\draw (11.85,-12.13)-- (11.85,-14.13);
\draw [color=zzttqq] (8.71,-15.24)-- (10.78,-13.26);
\draw [color=zzttqq] (10.78,-13.26)-- (8.8,-11.18);
\draw [color=zzttqq] (8.8,-11.18)-- (6.72,-13.17);
\draw [color=zzttqq] (6.72,-13.17)-- (8.71,-15.24);
\draw (6.02,-19.01)-- (6.02,-21.01);
\draw (7.97,-19.06)-- (7.97,-21.06);
\draw (10.04,-19.09)-- (10.04,-21.09);
\draw [color=zzttqq] (6.98,-22.03)-- (8.99,-20.09);
\draw [color=zzttqq] (8.99,-20.09)-- (7.04,-18.08);
\draw [color=zzttqq] (7.04,-18.08)-- (5.03,-20.02);
\draw [color=zzttqq] (5.03,-20.02)-- (6.98,-22.03);
\draw [domain=2.94:24.74] plot(\x,{(-402.19-0*\x)/20});
\draw (11.92,-19.09)-- (11.92,-21.09);
\draw [color=zzttqq] (10.97,-22.09)-- (12.98,-20.14);
\draw [color=zzttqq] (12.98,-20.14)-- (11.03,-18.13);
\draw [color=zzttqq] (11.03,-18.13)-- (9.02,-20.08);
\draw [color=zzttqq] (9.02,-20.08)-- (10.97,-22.09);
\draw [color=zzttqq] (12.76,-15.18)-- (14.77,-13.23);
\draw [color=zzttqq] (14.77,-13.23)-- (12.83,-11.22);
\draw [color=zzttqq] (12.83,-11.22)-- (10.82,-13.17);
\draw [color=zzttqq] (10.82,-13.17)-- (12.76,-15.18);
\draw [color=zzttqq] (16.73,-15.24)-- (18.74,-13.3);
\draw [color=zzttqq] (18.74,-13.3)-- (16.8,-11.29);
\draw [color=zzttqq] (16.8,-11.29)-- (14.79,-13.23);
\draw [color=zzttqq] (14.79,-13.23)-- (16.73,-15.24);
\draw [color=zzttqq] (14.95,-22.08)-- (16.96,-20.13);
\draw [color=zzttqq] (16.96,-20.13)-- (15.01,-18.12);
\draw [color=zzttqq] (15.01,-18.12)-- (13,-20.07);
\draw [color=zzttqq] (13,-20.07)-- (14.95,-22.08);
\draw [color=zzttqq] (20.76,-15.27)-- (22.77,-13.33);
\draw [color=zzttqq] (22.77,-13.33)-- (20.82,-11.32);
\draw [color=zzttqq] (20.82,-11.32)-- (18.81,-13.26);
\draw [color=zzttqq] (18.81,-13.26)-- (20.76,-15.27);
\draw [color=zzttqq] (18.9,-22.15)-- (20.91,-20.21);
\draw [color=zzttqq] (20.91,-20.21)-- (18.96,-18.2);
\draw [color=zzttqq] (18.96,-18.2)-- (16.95,-20.14);
\draw [color=zzttqq] (16.95,-20.14)-- (18.9,-22.15);
\draw [color=zzttqq] (24.77,-15.27)-- (26.78,-13.33);
\draw [color=zzttqq] (26.78,-13.33)-- (24.84,-11.32);
\draw [color=zzttqq] (24.84,-11.32)-- (22.83,-13.26);
\draw [color=zzttqq] (22.83,-13.26)-- (24.77,-15.27);
\draw [color=zzttqq] (3.04,-22)-- (5.05,-20.06);
\draw [color=zzttqq] (5.05,-20.06)-- (3.11,-18.05);
\draw [color=zzttqq] (3.11,-18.05)-- (1.1,-19.99);
\draw [color=zzttqq] (1.1,-19.99)-- (3.04,-22);
\draw [color=zzttqq] (22.83,-22.21)-- (24.84,-20.27);
\draw [color=zzttqq] (24.84,-20.27)-- (22.89,-18.26);
\draw [color=zzttqq] (22.89,-18.26)-- (20.88,-20.2);
\draw [color=zzttqq] (20.88,-20.2)-- (22.83,-22.21);
\draw (14.05,-19.04)-- (14.05,-21.04);
\draw (16.06,-19.04)-- (16.06,-21.04);
\draw (18,-19.1)-- (18,-21.1);
\draw (20.07,-19.12)-- (20.07,-21.12);
\draw (21.95,-19.12)-- (21.95,-21.12);
\draw (13.75,-12.19)-- (13.75,-14.19);
\draw (15.75,-12.19)-- (15.75,-14.19);
\draw (17.7,-12.25)-- (17.7,-14.25);
\draw (19.76,-12.27)-- (19.76,-14.27);
\draw (21.64,-12.27)-- (21.64,-14.27);
\begin{scriptsize}
\fill [color=qqqqff] (-10,17.32) circle (1.5pt);
\draw[color=qqqqff] (0.15,-2.87) node {$D$};
\fill [color=qqqqff] (2.8,-13.21) circle (1.5pt);
\draw[color=qqqqff] (3.25,-12.82) node {$C_2$};

\fill [color=qqqqff] (22.8,-13.21) circle (1.5pt);
\draw[color=qqqqff] (23.24,-12.82) node {$B_1$};

\fill [color=qqqqff] (24.74,-13.21) circle (1.5pt);
\draw[color=qqqqff] (25.19,-12.82) node {$B_4$};

\fill [color=qqqqff] (5.79,-12.17) circle (1.5pt);
\draw[color=qqqqff] (6.23,-11.79) node {$E_3$};
\fill [color=qqqqff] (5.79,-14.17) circle (1.5pt);
\draw[color=qqqqff] (6.23,-13.8) node {$G_3$};
\fill [color=qqqqff] (7.85,-12.22) circle (1.5pt);
\draw[color=qqqqff] (8.3,-11.85) node {$E_5$};
\fill [color=qqqqff] (7.85,-14.22) circle (1.5pt);
\draw[color=qqqqff] (8.3,-13.82) node {$G_5$};
\fill [color=qqqqff] (9.74,-12.11) circle (1.5pt);
\draw[color=qqqqff] (10.19,-11.73) node {$E_4$};
\fill [color=qqqqff] (9.74,-14.11) circle (1.5pt);
\draw[color=qqqqff] (10.19,-13.74) node {$G_4$};
\fill [color=qqqqff] (4.74,-15.17) circle (1.5pt);
\draw[color=qqqqff] (5.14,-14.77) node {$F_2$};
\fill [color=uququq] (4.8,-11.21) circle (1.5pt);
\draw[color=uququq] (5.17,-10.81) node {$J_1$};
\fill [color=uququq] (2.79,-13.16) circle (1.5pt);
\draw[color=uququq] (3.22,-12.76) node {$K_1$};
\fill [color=qqqqff] (11.85,-12.13) circle (1.5pt);
\draw[color=qqqqff] (12.28,-11.76) node {$E_1$};
\fill [color=qqqqff] (11.85,-14.13) circle (1.5pt);
\draw[color=qqqqff] (12.28,-13.74) node {$G_1$};
\fill [color=qqqqff] (8.71,-15.24) circle (1.5pt);
\draw[color=qqqqff] (9.13,-14.86) node {$F_3$};
\fill [color=uququq] (8.8,-11.18) circle (1.5pt);
\draw[color=uququq] (9.15,-10.81) node {$J_2$};
\fill [color=uququq] (6.72,-13.17) circle (1.5pt);
\draw[color=uququq] (7.15,-12.79) node {$K_2$};
\fill [color=qqqqff] (-9.77,10.48) circle (1.5pt);
\draw[color=qqqqff] (0.36,-2.87) node {$D_2$};
\fill [color=qqqqff] (2.94,-20.11) circle (1.5pt);
\draw[color=qqqqff] (3.4,-19.73) node {$C_4$};
\fill [color=qqqqff] (22.94,-20.11) circle (1.5pt);
\draw[color=qqqqff] (23.39,-19.97) node {$B_2$};

\fill [color=qqqqff] (24.74,-20.11) circle (1.5pt);
\draw[color=qqqqff] (25.19,-19.97) node {$B_3$};

\fill [color=qqqqff] (6.02,-19.01) circle (1.5pt);
\draw[color=qqqqff] (6.47,-18.64) node {$E_8$};
\fill [color=qqqqff] (6.02,-21.01) circle (1.5pt);
\draw[color=qqqqff] (6.47,-20.62) node {$G_8$};
\fill [color=qqqqff] (7.97,-19.06) circle (1.5pt);
\draw[color=qqqqff] (8.77,-18.67) node {$E_{10}$};
\fill [color=qqqqff] (7.97,-21.06) circle (1.5pt);
\draw[color=qqqqff] (8.77,-20.68) node {$G_{10}$};
\fill [color=qqqqff] (10.04,-19.09) circle (1.5pt);
\draw[color=qqqqff] (10.48,-18.7) node {$E_9$};
\fill [color=qqqqff] (10.04,-21.09) circle (1.5pt);
\draw[color=qqqqff] (10.48,-20.7) node {$G_9$};
\fill [color=qqqqff] (6.98,-22.03) circle (1.5pt);
\draw[color=qqqqff] (7.38,-21.65) node {$F_5$};
\fill [color=uququq] (7.04,-18.08) circle (1.5pt);
\draw[color=uququq] (7.38,-17.69) node {$J_3$};
\fill [color=uququq] (5.03,-20.02) circle (1.5pt);
\draw[color=uququq] (5.43,-19.64) node {$K_3$};
\fill [color=qqqqff] (11.92,-19.09) circle (1.5pt);
\draw[color=qqqqff] (12.37,-18.7) node {$E_6$};
\fill [color=qqqqff] (11.92,-21.09) circle (1.5pt);
\draw[color=qqqqff] (12.37,-20.7) node {$G_6$};
\fill [color=qqqqff] (10.97,-22.09) circle (1.5pt);
\draw[color=qqqqff] (11.37,-21.71) node {$F_6$};
\fill [color=uququq] (11.03,-18.13) circle (1.5pt);
\draw[color=uququq] (11.4,-17.75) node {$J_4$};
\fill [color=uququq] (9.02,-20.08) circle (1.5pt);
\draw[color=uququq] (9.45,-19.7) node {$K_4$};
\fill [color=qqqqff] (12.76,-15.18) circle (1.5pt);
\draw[color=qqqqff] (13.17,-14.8) node {$F_7$};
\fill [color=uququq] (12.83,-11.22) circle (1.5pt);
\draw[color=uququq] (13.17,-10.84) node {$J_5$};
\fill [color=uququq] (10.82,-13.17) circle (1.5pt);
\draw[color=uququq] (11.22,-12.79) node {$K_5$};
\fill [color=qqqqff] (16.73,-15.24) circle (1.5pt);
\draw[color=qqqqff] (17.16,-14.86) node {$F_8$};
\fill [color=uququq] (16.8,-11.29) circle (1.5pt);
\draw[color=uququq] (17.16,-10.9) node {$J_6$};
\fill [color=uququq] (14.79,-13.23) circle (1.5pt);
\draw[color=uququq] (15.21,-12.85) node {$K_6$};
\fill [color=qqqqff] (14.95,-22.08) circle (1.5pt);
\draw[color=qqqqff] (15.36,-21.68) node {$F_9$};
\fill [color=uququq] (15.01,-18.12) circle (1.5pt);
\draw[color=uququq] (15.36,-17.72) node {$J_7$};
\fill [color=uququq] (13,-20.07) circle (1.5pt);
\draw[color=uququq] (13.41,-19.67) node {$K_7$};
\fill [color=qqqqff] (20.76,-15.27) circle (1.5pt);
\draw[color=qqqqff] (21.53,-14.89) node {$F_{10}$};
\fill [color=uququq] (20.82,-11.32) circle (1.5pt);
\draw[color=uququq] (21.17,-10.93) node {$J_8$};
\fill [color=uququq] (18.81,-13.26) circle (1.5pt);
\draw[color=uququq] (19.22,-12.88) node {$K_8$};
\fill [color=qqqqff] (18.9,-22.15) circle (1.5pt);
\draw[color=qqqqff] (19.67,-21.77) node {$F_{11}$};
\fill [color=uququq] (18.96,-18.2) circle (1.5pt);
\draw[color=uququq] (19.31,-17.81) node {$J_9$};
\fill [color=uququq] (16.95,-20.14) circle (1.5pt);
\draw[color=uququq] (17.36,-19.76) node {$K_9$};
\fill [color=qqqqff] (24.77,-15.27) circle (1.5pt);
\draw[color=qqqqff] (25.54,-14.89) node {$F_{12}$};
\fill [color=uququq] (24.84,-11.32) circle (1.5pt);
\draw[color=uququq] (25.54,-10.93) node {$J_{10}$};
\fill [color=uququq] (22.83,-13.26) circle (1.5pt);
\draw[color=uququq] (23.59,-12.88) node {$K_{10}$};
\fill [color=qqqqff] (3.04,-22) circle (1.5pt);
\draw[color=qqqqff] (3.81,-21.62) node {$F_{13}$};
\fill [color=uququq] (3.11,-18.05) circle (1.5pt);
\draw[color=uququq] (3.81,-17.66) node {$J_{11}$};
\fill [color=uququq] (1.1,-19.99) circle (1.5pt);
\draw[color=uququq] (1.86,-19.61) node {$K_{11}$};
\fill [color=qqqqff] (22.83,-22.21) circle (1.5pt);
\draw[color=qqqqff] (23.59,-21.83) node {$F_{14}$};
\fill [color=uququq] (22.89,-18.26) circle (1.5pt);
\draw[color=uququq] (23.59,-17.87) node {$J_{12}$};
\fill [color=uququq] (20.88,-20.2) circle (1.5pt);
\draw[color=uququq] (21.64,-19.82) node {$K_{12}$};
\fill [color=qqqqff] (14.05,-19.04) circle (1.5pt);
\draw[color=qqqqff] (14.85,-18.67) node {$E_{11}$};
\fill [color=qqqqff] (14.05,-21.04) circle (1.5pt);
\draw[color=qqqqff] (14.85,-20.65) node {$G_{11}$};
\fill [color=qqqqff] (16.06,-19.04) circle (1.5pt);
\draw[color=qqqqff] (16.86,-18.67) node {$E_{12}$};
\fill [color=qqqqff] (16.06,-21.04) circle (1.5pt);
\draw[color=qqqqff] (16.86,-20.65) node {$G_{12}$};
\fill [color=qqqqff] (18,-19.1) circle (1.5pt);
\draw[color=qqqqff] (18.81,-18.73) node {$E_{13}$};
\fill [color=qqqqff] (18,-21.1) circle (1.5pt);
\draw[color=qqqqff] (18.81,-20.7) node {$G_{13}$};
\fill [color=qqqqff] (20.07,-19.12) circle (1.5pt);
\draw[color=qqqqff] (20.88,-18.73) node {$E_{14}$};
\fill [color=qqqqff] (20.07,-21.12) circle (1.5pt);
\draw[color=qqqqff] (20.88,-20.73) node {$G_{14}$};
\fill [color=uququq] (15.06,-20.05) circle (1.5pt);
\draw[color=uququq] (15.83,-19.67) node {$K_{13}$};
\fill [color=qqqqff] (21.95,-19.12) circle (1.5pt);
\draw[color=qqqqff] (22.74,-18.73) node {$E_{15}$};
\fill [color=qqqqff] (21.95,-21.12) circle (1.5pt);
\draw[color=qqqqff] (22.74,-20.73) node {$G_{15}$};
\fill [color=uququq] (19.06,-20.11) circle (1.5pt);
\draw[color=uququq] (19.81,-19.73) node {$K_{14}$};
\fill [color=qqqqff] (12.67,-13.29) circle (1.5pt);
\draw[color=qqqqff] (13.11,-12.91) node {$C_6$};
\fill [color=qqqqff] (13.75,-12.19) circle (1.5pt);
\draw[color=qqqqff] (14.53,-11.82) node {$E_{16}$};
\fill [color=qqqqff] (13.75,-14.19) circle (1.5pt);
\draw[color=qqqqff] (14.53,-13.8) node {$G_{16}$};
\fill [color=qqqqff] (15.75,-12.19) circle (1.5pt);
\draw[color=qqqqff] (16.54,-11.82) node {$E_{17}$};
\fill [color=qqqqff] (15.75,-14.19) circle (1.5pt);
\draw[color=qqqqff] (16.54,-13.8) node {$G_{17}$};
\fill [color=qqqqff] (17.7,-12.25) circle (1.5pt);
\draw[color=qqqqff] (18.49,-11.88) node {$E_{18}$};
\fill [color=qqqqff] (17.7,-14.25) circle (1.5pt);
\draw[color=qqqqff] (18.49,-13.85) node {$G_{18}$};
\fill [color=qqqqff] (19.76,-12.27) circle (1.5pt);
\draw[color=qqqqff] (20.55,-11.88) node {$E_{19}$};
\fill [color=qqqqff] (19.76,-14.27) circle (1.5pt);
\draw[color=qqqqff] (20.55,-13.88) node {$G_{19}$};
\fill [color=uququq] (14.76,-13.21) circle (1.5pt);
\draw[color=uququq] (15.53,-12.82) node {$K_{15}$};
\fill [color=qqqqff] (21.64,-12.27) circle (1.5pt);
\draw[color=qqqqff] (22.44,-11.88) node {$E_{20}$};
\fill [color=qqqqff] (21.64,-14.27) circle (1.5pt);
\draw[color=qqqqff] (22.44,-13.88) node {$G_{20}$};
\fill [color=uququq] (18.75,-13.26) circle (1.5pt);
\draw[color=uququq] (19.52,-12.88) node {$K_{16}$};
\end{scriptsize}
\end{tikzpicture}}
    \caption{Structure within the odd length path, where each unit length segment is now shrunk to $\sqrt{2}/2$, and the vertical
    bar length is $\zeta = \sqrt{2}/2$.}
    \label{fig:rectangular_edge_pattern}
\end{figure}

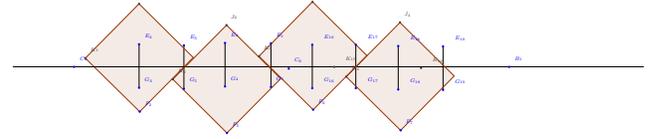
\begin{figure}[ht]
    \definecolor{uququq}{rgb}{0.25,0.25,0.25}
\definecolor{zzttqq}{rgb}{0.6,0.2,0}
\definecolor{qqqqff}{rgb}{0,0,1}
\centering
\resizebox{\linewidth}{!}{

\begin{tikzpicture}[line cap=round,line join=round,>=triangle 45,x=1.0cm,y=1.0cm]
\clip(0.03,-23.18) rectangle (28.94,-10.08);
\fill[color=zzttqq,fill=zzttqq,fill opacity=0.1] (5.82,-15.27) -- (8.28,-12.78) -- (5.79,-10.31) -- (3.32,-12.81) -- cycle;
\fill[color=zzttqq,fill=zzttqq,fill opacity=0.1] (9.83,-16.25) -- (12.3,-13.75) -- (9.81,-11.29) -- (7.34,-13.78) -- cycle;
\fill[color=zzttqq,fill=zzttqq,fill opacity=0.1] (13.79,-15.18) -- (16.26,-12.69) -- (13.76,-10.22) -- (11.3,-12.72) -- cycle;
\fill[color=zzttqq,fill=zzttqq,fill opacity=0.1] (17.81,-16.13) -- (20.27,-13.63) -- (17.78,-11.17) -- (15.31,-13.66) -- cycle;
\draw (5.79,-12.17)-- (5.79,-14.17);
\draw (7.85,-12.22)-- (7.85,-14.22);
\draw (9.74,-12.11)-- (9.74,-14.11);
\draw [color=zzttqq] (5.82,-15.27)-- (8.28,-12.78);
\draw [color=zzttqq] (8.28,-12.78)-- (5.79,-10.31);
\draw [color=zzttqq] (5.79,-10.31)-- (3.32,-12.81);
\draw [color=zzttqq] (3.32,-12.81)-- (5.82,-15.27);
\draw [domain=-1.21:35.7] plot(\x,{(-264.11-0*\x)/20});
\draw (11.85,-12.13)-- (11.85,-14.13);
\draw (13.75,-12.19)-- (13.75,-14.19);
\draw (15.75,-12.19)-- (15.75,-14.19);
\draw (17.7,-12.25)-- (17.7,-14.25);
\draw (19.76,-12.27)-- (19.76,-14.27);
\draw [color=zzttqq] (9.83,-16.25)-- (12.3,-13.75);
\draw [color=zzttqq] (12.3,-13.75)-- (9.81,-11.29);
\draw [color=zzttqq] (9.81,-11.29)-- (7.34,-13.78);
\draw [color=zzttqq] (7.34,-13.78)-- (9.83,-16.25);
\draw [color=zzttqq] (13.79,-15.18)-- (16.26,-12.69);
\draw [color=zzttqq] (16.26,-12.69)-- (13.76,-10.22);
\draw [color=zzttqq] (13.76,-10.22)-- (11.3,-12.72);
\draw [color=zzttqq] (11.3,-12.72)-- (13.79,-15.18);
\draw [color=zzttqq] (17.81,-16.13)-- (20.27,-13.63);
\draw [color=zzttqq] (20.27,-13.63)-- (17.78,-11.17);
\draw [color=zzttqq] (17.78,-11.17)-- (15.31,-13.66);
\draw [color=zzttqq] (15.31,-13.66)-- (17.81,-16.13);
\begin{scriptsize}
\fill [color=qqqqff] (-10,17.32) circle (1.5pt);
\draw[color=qqqqff] (-1.09,-3.28) node {$D$};
\fill [color=qqqqff] (2.8,-13.21) circle (1.5pt);
\draw[color=qqqqff] (3.25,-12.82) node {$C_2$};
\fill [color=qqqqff] (22.8,-13.21) circle (1.5pt);
\draw[color=qqqqff] (23.24,-12.82) node {$B_1$};
\fill [color=qqqqff] (5.79,-12.17) circle (1.5pt);
\draw[color=qqqqff] (6.23,-11.79) node {$E_3$};
\fill [color=qqqqff] (5.79,-14.17) circle (1.5pt);
\draw[color=qqqqff] (6.23,-13.8) node {$G_3$};
\fill [color=qqqqff] (7.85,-12.22) circle (1.5pt);
\draw[color=qqqqff] (8.3,-11.85) node {$E_5$};
\fill [color=qqqqff] (7.85,-14.22) circle (1.5pt);
\draw[color=qqqqff] (8.3,-13.82) node {$G_5$};
\fill [color=qqqqff] (9.74,-12.11) circle (1.5pt);
\draw[color=qqqqff] (10.19,-11.73) node {$E_4$};
\fill [color=qqqqff] (9.74,-14.11) circle (1.5pt);
\draw[color=qqqqff] (10.19,-13.74) node {$G_4$};
\fill [color=qqqqff] (5.82,-15.27) circle (1.5pt);
\draw[color=qqqqff] (6.23,-14.89) node {$F_2$};
\fill [color=uququq] (5.79,-10.31) circle (1.5pt);
\draw[color=uququq] (6.14,-9.93) node {$J_1$};
\fill [color=uququq] (3.32,-12.81) circle (1.5pt);
\draw[color=uququq] (3.75,-12.41) node {$K_1$};
\fill [color=qqqqff] (11.85,-12.13) circle (1.5pt);
\draw[color=qqqqff] (12.28,-11.76) node {$E_1$};
\fill [color=qqqqff] (11.85,-14.13) circle (1.5pt);
\draw[color=qqqqff] (12.28,-13.74) node {$G_1$};
\fill [color=qqqqff] (-9.77,10.48) circle (1.5pt);
\draw[color=qqqqff] (-0.88,-3.28) node {$D_2$};
\fill [color=qqqqff] (12.67,-13.29) circle (1.5pt);
\draw[color=qqqqff] (13.11,-12.91) node {$C_6$};
\fill [color=qqqqff] (13.75,-12.19) circle (1.5pt);
\draw[color=qqqqff] (14.53,-11.82) node {$E_{16}$};
\fill [color=qqqqff] (13.75,-14.19) circle (1.5pt);
\draw[color=qqqqff] (14.53,-13.8) node {$G_{16}$};
\fill [color=qqqqff] (15.75,-12.19) circle (1.5pt);
\draw[color=qqqqff] (16.54,-11.82) node {$E_{17}$};
\fill [color=qqqqff] (15.75,-14.19) circle (1.5pt);
\draw[color=qqqqff] (16.54,-13.8) node {$G_{17}$};
\fill [color=qqqqff] (17.7,-12.25) circle (1.5pt);
\draw[color=qqqqff] (18.49,-11.88) node {$E_{18}$};
\fill [color=qqqqff] (17.7,-14.25) circle (1.5pt);
\draw[color=qqqqff] (18.49,-13.85) node {$G_{18}$};
\fill [color=qqqqff] (19.76,-12.27) circle (1.5pt);
\draw[color=qqqqff] (20.55,-11.88) node {$E_{19}$};
\fill [color=qqqqff] (19.76,-14.27) circle (1.5pt);
\draw[color=qqqqff] (20.55,-13.88) node {$G_{19}$};
\fill [color=uququq] (14.76,-13.21) circle (1.5pt);
\draw[color=uququq] (15.53,-12.82) node {$K_{15}$};
\fill [color=uququq] (18.75,-13.26) circle (1.5pt);
\draw[color=uququq] (19.52,-12.88) node {$K_{16}$};
\fill [color=qqqqff] (9.83,-16.25) circle (1.5pt);
\draw[color=qqqqff] (10.25,-15.86) node {$F_5$};
\fill [color=uququq] (9.81,-11.29) circle (1.5pt);
\draw[color=uququq] (10.16,-10.9) node {$J_2$};
\fill [color=uququq] (7.34,-13.78) circle (1.5pt);
\draw[color=uququq] (7.77,-13.38) node {$K_2$};
\fill [color=qqqqff] (13.79,-15.18) circle (1.5pt);
\draw[color=qqqqff] (14.2,-14.8) node {$F_6$};
\fill [color=uququq] (13.76,-10.22) circle (1.5pt);
\draw[color=uququq] (14.12,-9.84) node {$J_3$};
\fill [color=uququq] (11.3,-12.72) circle (1.5pt);
\draw[color=uququq] (11.72,-12.32) node {$K_3$};
\fill [color=qqqqff] (17.81,-16.13) circle (1.5pt);
\draw[color=qqqqff] (18.22,-15.74) node {$F_7$};
\fill [color=uququq] (17.78,-11.17) circle (1.5pt);
\draw[color=uququq] (18.13,-10.78) node {$J_4$};
\fill [color=uququq] (15.31,-13.66) circle (1.5pt);
\draw[color=uququq] (15.74,-13.26) node {$K_4$};
\end{scriptsize}
\end{tikzpicture}
}
\caption{Possible square coverage pattern along the constructed odd-length path 
    until the square side length $\ell$ reaches $1.25$.}
\label{fig:extreme_box_coverage_pattern}
\end{figure}

In this way, the pattern for covering a previous odd-length path in the structural gadget 
should be one of the two shown in Figure~\ref{fig:rectangular_edge_pattern} until the 
square side length $\ell$ reaches $1.25$, and the pattern in Figure~\ref{fig:extreme_box_coverage_pattern} may appear.

At the junction of the three paths, a tri-connected crossing is made where the distance between the 
crossing point and the neighboring vertical bar is set to be $3\sqrt{2}/4$.

\begin{figure}[ht]
\centering
\definecolor{uququq}{rgb}{0.25,0.25,0.25}
\definecolor{zzttqq}{rgb}{0.6,0.2,0}
\definecolor{qqqqff}{rgb}{0,0,1}
\resizebox{\linewidth}{!}{
\begin{tikzpicture}[line cap=round,line join=round,>=triangle 45,x=1.0cm,y=1.0cm]
\clip(-6.72,-6.03) rectangle (9.8,6.04);
\fill[color=zzttqq,fill=zzttqq,fill opacity=0.1] (1.95,-2.08) -- (3.97,-0.07) -- (1.96,1.95) -- (-0.06,-0.06) -- cycle;
\fill[color=zzttqq,fill=zzttqq,fill opacity=0.1] (-2.76,0.74) -- (-1.94,3.47) -- (0.79,2.65) -- (-0.03,-0.08) -- cycle;
\fill[color=zzttqq,fill=zzttqq,fill opacity=0.1] (-2.07,-3.4) -- (0.66,-2.73) -- (-0.01,-0.01) -- (-2.73,-0.67) -- cycle;
\draw [domain=-6.721631532290574:-0.008222052616553954] plot(\x,{(--0.12--8.98*\x)/-4.99});
\draw [domain=-6.721631532290574:-0.008222052616553954] plot(\x,{(-0.03-10.48*\x)/-6.27});
\draw [domain=-0.008222052616553954:9.803628968516454] plot(\x,{(-0.1-0.24*\x)/10.82});
\draw [color=zzttqq] (1.95,-2.08)-- (3.97,-0.07);
\draw [color=zzttqq] (3.97,-0.07)-- (1.96,1.95);
\draw [color=zzttqq] (1.96,1.95)-- (-0.06,-0.06);
\draw [color=zzttqq] (-0.06,-0.06)-- (1.95,-2.08);
\draw (3,1)-- (3,-1);
\draw (5,1)-- (5,-1);
\draw (-0.54,3.12)-- (-2.31,2.17);
\draw (-1.49,4.88)-- (-3.25,3.94);
\draw (-2.45,-2.03)-- (-0.76,-3.09);
\draw (-3.51,-3.73)-- (-1.81,-4.79);

\draw [color=zzttqq] (-2.76,0.74) -- (-1.94,3.47);
\draw [color=zzttqq] (-1.94,3.47) -- (0.79,2.65);
\draw [color=zzttqq] (0.79,2.65) -- (-0.03,-0.08);
\draw [color=zzttqq] (-0.03,-0.08)-- (-2.76,0.74);

\draw [color=zzttqq] (-2.07,-3.4)-- (0.66,-2.73);
\draw [color=zzttqq] (0.66,-2.73) -- (-0.01,-0.01);
\draw [color=zzttqq] (-0.01,-0.01)-- (-2.73,-0.67);
\draw [color=zzttqq] (-2.73,-0.67)-- (-2.07,-3.4);

\begin{scriptsize}
\fill [color=qqqqff] (-0.01,-0.01) circle (1.5pt);
\draw[color=qqqqff] (0.14,0.23) node {$C$};
\fill [color=qqqqff] (1.95,-2.08) circle (1.5pt);
\draw[color=qqqqff] (2.08,-1.84) node {$F$};
\fill [color=uququq] (1.96,1.95) circle (1.5pt);
\draw[color=uququq] (2.04,2.17) node {$J$};
\fill [color=uququq] (-0.06,-0.06) circle (1.5pt);
\draw[color=uququq] (0.07,0.18) node {$K$};
\fill [color=qqqqff] (3,1) circle (1.5pt);
\draw[color=qqqqff] (3.14,1.24) node {$E$};
\fill [color=qqqqff] (3,-1) circle (1.5pt);
\draw[color=qqqqff] (3.14,-0.78) node {$G$};
\fill [color=qqqqff] (5,1) circle (1.5pt);
\draw[color=qqqqff] (5.26,1.24) node {$E_1$};
\fill [color=qqqqff] (5,-1) circle (1.5pt);
\draw[color=qqqqff] (5.26,-0.78) node {$G_1$};
\fill [color=qqqqff] (-0.54,3.12) circle (1.5pt);
\draw[color=qqqqff] (-0.38,3.34) node {$E'$};
\fill [color=qqqqff] (-2.31,2.17) circle (1.5pt);
\draw[color=qqqqff] (-2.13,2.4) node {$G'$};
\fill [color=qqqqff] (-1.49,4.88) circle (1.5pt);
\draw[color=qqqqff] (-1.35,5.11) node {$H$};
\fill [color=qqqqff] (-3.25,3.94) circle (1.5pt);
\draw[color=qqqqff] (-3.13,4.17) node {$L$};
\fill [color=qqqqff] (-2.45,-2.03) circle (1.5pt);
\draw[color=qqqqff] (-2.18,-1.8) node {$E'_1$};
\fill [color=qqqqff] (-0.76,-3.09) circle (1.5pt);
\draw[color=qqqqff] (-0.45,-2.86) node {$G'_1$};
\fill [color=qqqqff] (-3.51,-3.73) circle (1.5pt);
\draw[color=qqqqff] (-3.35,-3.5) node {$M$};
\fill [color=qqqqff] (-1.81,-4.79) circle (1.5pt);
\draw[color=qqqqff] (-1.67,-4.56) node {$N$};
\fill [color=uququq] (-0.01,-0.01) circle (1.5pt);
\draw[color=uququq] (0.14,0.23) node {$Q$};
\fill [color=uququq] (-2.73,-0.67) circle (1.5pt);
\draw[color=uququq] (-2.59,-0.44) node {$R$};
\end{scriptsize}
\end{tikzpicture}}

    \caption{Square coverage pattern at the junction crossing with three unit squares.}
    \label{fig:sq_crossing}
\end{figure}
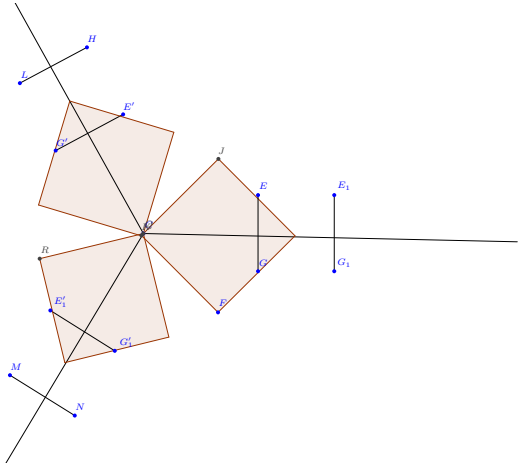

So, when there exists a vertex cover with $n$ vertices for Problem~\ref{p:vc_cubic}, we can use $n + M$
unit squares to cover all the vertices.

Then, let us prove that when there are $n+M$ squares with edge length less than $1.165$ covering the tri-net structure, there is a vertex cover of size $n$ 
for the graph $G$.

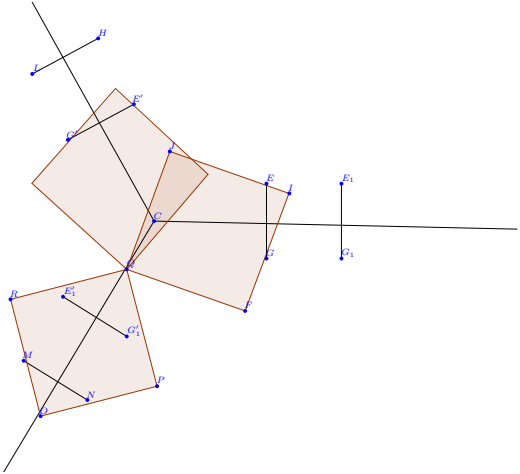
\begin{figure}[ht]
\definecolor{uququq}{rgb}{0.25,0.25,0.25}
\definecolor{zzttqq}{rgb}{0.6,0.2,0}
\definecolor{qqqqff}{rgb}{0,0,1}
\centering
\resizebox{\linewidth}{!}{%
\begin{tikzpicture}[line cap=round,line join=round,x=1.0cm,y=1.0cm]
\clip(-7.2,-6.69) rectangle (9.69,5.84);
\fill[color=zzttqq,fill=zzttqq,fill opacity=0.1] (2.43,-2.4) -- (3.60,0.74) -- (0.42,1.86) -- (-0.73,-1.29) -- cycle;
\fill[color=zzttqq,fill=zzttqq,fill opacity=0.1] (-3.26,1.01) -- (-1.03,3.48) -- (1.44,1.25) -- (-0.73,-1.29) -- cycle;
\fill[color=zzttqq,fill=zzttqq,fill opacity=0.1] (-3.03,-5.21) -- (0.08,-4.41) -- (-0.73,-1.29) -- (-3.83,-2.09) -- cycle;
\draw [domain=-7.198708911238607:0.0] plot(\x,{(-0--8.97*\x)/-4.99});
\draw [domain=-7.198708911238607:0.0] plot(\x,{(-0-10.49*\x)/-6.28});
\draw [domain=0.0:9.687137709537886] plot(\x,{(-0-0.24*\x)/10.81});
\draw [color=zzttqq] (2.43,-2.4)-- (3.60,0.74);
\draw [color=zzttqq] (3.60,0.74)-- (0.42,1.86);
\draw [color=zzttqq] (0.42,1.86)-- (-0.73,-1.29);
\draw [color=zzttqq] (-0.73,-1.29)-- (2.43,-2.4);
\draw (3,1)-- (3,-1);
\draw (5,1)-- (5,-1);
\draw (-0.54,3.12)-- (-2.3,2.17);
\draw (-1.49,4.88)-- (-3.25,3.93);
\draw (-2.43,-2.02)-- (-0.73,-3.08);
\draw (-3.48,-3.73)-- (-1.78,-4.78);
\draw [color=zzttqq] (-3.03,-5.21)-- (0.08,-4.41);
\draw [color=zzttqq] (0.08,-4.41)-- (-0.73,-1.29);
\draw [color=zzttqq] (-0.73,-1.29)-- (-3.83,-2.09);
\draw [color=zzttqq] (-3.83,-2.09)-- (-3.03,-5.21);

\draw [color=zzttqq] (-3.26,1.01) -- (-1.03,3.54);
\draw [color=zzttqq] (-1.03,3.54) -- (1.44,1.25);
\draw [color=zzttqq] (1.44,1.25) -- (-0.73,-1.29);
\draw [color=zzttqq] (-0.73,-1.29) -- (-3.26,1.01);

\begin{scriptsize}
\fill [color=qqqqff] (0,0) circle (1.5pt);
\draw[color=qqqqff] (0.1,0.15) node {$C$};
\fill [color=qqqqff] (2.43,-2.4) circle (1.5pt);
\draw[color=qqqqff] (2.52,-2.25) node {$F$};
\fill [color=qqqqff] (3.61,0.74) circle (1.5pt);
\draw[color=qqqqff] (3.65,0.89) node {$I$};
\fill [color=qqqqff] (0.42,1.86) circle (1.5pt);
\draw[color=qqqqff] (0.48,2.02) node {$J$};


\fill [color=qqqqff] (3,1) circle (1.5pt);
\draw[color=qqqqff] (3.1,1.15) node {$E$};
\fill [color=qqqqff] (3,-1) circle (1.5pt);
\draw[color=qqqqff] (3.1,-0.84) node {$G$};
\fill [color=qqqqff] (5,1) circle (1.5pt);
\draw[color=qqqqff] (5.18,1.15) node {$E_1$};
\fill [color=qqqqff] (5,-1) circle (1.5pt);
\draw[color=qqqqff] (5.18,-0.85) node {$G_1$};
\fill [color=qqqqff] (-0.54,3.12) circle (1.5pt);
\draw[color=qqqqff] (-0.43,3.27) node {$E'$};
\fill [color=qqqqff] (-2.3,2.17) circle (1.5pt);
\draw[color=qqqqff] (-2.18,2.32) node {$G'$};
\fill [color=qqqqff] (-1.49,4.88) circle (1.5pt);
\draw[color=qqqqff] (-1.37,5.03) node {$H$};
\fill [color=qqqqff] (-3.25,3.93) circle (1.5pt);
\draw[color=qqqqff] (-3.13,4.08) node {$L$};
\fill [color=qqqqff] (-2.43,-2.02) circle (1.5pt);
\draw[color=qqqqff] (-2.24,-1.87) node {$E'_1$};
\fill [color=qqqqff] (-0.73,-3.08) circle (1.5pt);
\draw[color=qqqqff] (-0.53,-2.92) node {$G'_1$};
\fill [color=qqqqff] (-3.48,-3.73) circle (1.5pt);
\draw[color=qqqqff] (-3.37,-3.57) node {$M$};
\fill [color=qqqqff] (-1.78,-4.78) circle (1.5pt);
\draw[color=qqqqff] (-1.68,-4.63) node {$N$};
\fill [color=qqqqff] (-3.03,-5.21) circle (1.5pt);
\draw[color=qqqqff] (-2.94,-5.06) node {$O$};
\fill [color=qqqqff] (0.08,-4.41) circle (1.5pt);
\draw[color=qqqqff] (0.18,-4.25) node {$P$};
\fill [color=qqqqff] (-0.73,-1.29) circle (1.5pt);
\draw[color=qqqqff] (-0.62,-1.14) node {$Q$};
\fill [color=qqqqff] (-3.83,-2.09) circle (1.5pt);
\draw[color=qqqqff] (-3.74,-1.94) node {$R$};
\end{scriptsize}
\end{tikzpicture}}
    \caption{The pattern of covering the crossing in the special structure constructed at the extreme square side length $\ell$.}
    \label{fig:crossing}
\end{figure}

In the crossing, for 3 squares to cover all neighboring parts of $T$ using the pattern in Figure~\ref{fig:crossing},
the square side length $\ell$ must follow the pattern in Figure~\ref{fig:crossing}.

At the maximum square side length $\ell$ such that the structure
still needs the same number of squares to cover as when using unit
squares, Figure~\ref{fig:crossing} shows this situation when the junction can
be covered with three squares, among which two squares cover one vertical bar each and
one square covers two vertical bars. 

Listing all necessary geometric constraints in Figure~\ref{fig:crossing} gives
\begin{align}
    \label{eq:crossing_start}
    \|OQ\| + \|CQ\| &= \sqrt{2} \ell + \|CQ\| = 3\sqrt{2} / 2\\
    \|QG\|^2 &= (\frac{\|CQ\|}{2} + \frac{3\sqrt{2}}{4})^2 + ( \frac{\sqrt{3}}{2}\|CQ\| - \frac{\zeta}{2})^2 \\
    \|QG\|^2 &= \ell^2 + \|FG\|^2\\
    \|QE\|^2 &= (\frac{\|CQ\|}{2} + \frac{3\sqrt{2}}{4})^2 + ( \frac{\sqrt{3}}{2}\|CQ\| + \frac{\zeta}{2})^2 \\
    \|QE\|^2 &= \ell^2 + \|JE\|^2\\
    \|IE\|^2 + \|IG\|^2 &= \|EG\|^2 = \zeta^2 = 0.5 \\
    \|IG\| + \|GF\| &= \|IE\| + \|EJ\| = \ell
    \label{eq:crossing_end}
\end{align}

Similarly, solving the system of equations in \eqref{eq:crossing_start} to 
\eqref{eq:crossing_end} gives $\ell \approx 1.165$.

Thus, if we are asked to place $k$ squares to cover the boundary of a simple polygon, it is computationally intractable
to approximate the smallest side length of the square to cover the whole structure within a factor of $1.165$.

\begin{theorem}
    It is NP-hard to approximate Problem~\ref{p:sq_cov} within a factor of $1.165$.
\end{theorem}

This inapprox.\ factor can be interpreted as the maximum zoom factor the gimbal camera
can have before taking pictures, or as relating to the minimum Ground Sampling Distance (GSD) 
that $k$ pictures can provide subject to full coverage of the boundary of a simple region using a polynomial-time
algorithm unless $\mathcal{P}=\mathcal{NP}$.

\section{Hardness of Approximate Coverage with Restricted Locations Inside the Region}
In the construction and also in the problem setting in Problems~\ref{p:cir_cov} and~\ref{p:sq_cov}, 
it is assumed that the center of the circle or square can be anywhere on the plane.
However, in many robotics scenarios, the centers of the circles or squares, i.e., the robot locations, 
have restrictions; for example, they must be on top of the perimeter or inside the region.
If those restrictions are applied, are these problems still NP-hard, or 
can they be proved to have a higher inapproximability ratio?

Since if there is a solution of $k$ unit circles to cover the simple polygon constructed, 
they can be made to be centered at the skeleton structure $T_G$ constructed before
until the circle radius $\ell$ reaches $1.152$, 
the restricted version of the problem has at least the same inapproximability ratio.

So, for the restricted circle coverage problem, we have
\begin{theorem}
    It is NP-hard to approximate the circle version of Problem~\ref{p:con_sq_cir_cov} within a factor of $1.152$ when restricting
    the circle centers to the region boundary.
\end{theorem}

While for the restricted square coverage problem, a finer analysis of the crossing part at the junction gives the following. 

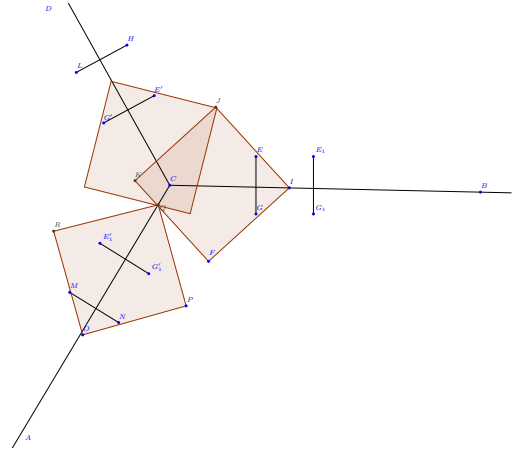
\begin{figure}[H]

\definecolor{uququq}{rgb}{0.25,0.25,0.25}
\definecolor{zzttqq}{rgb}{0.6,0.2,0}
\definecolor{qqqqff}{rgb}{0,0,1}
\centering
\resizebox{\linewidth}{!}{%
\definecolor{uququq}{rgb}{0.25,0.25,0.25}
\definecolor{zzttqq}{rgb}{0.6,0.2,0}
\definecolor{qqqqff}{rgb}{0,0,1}
\begin{tikzpicture}[line cap=round,line join=round,x=1.0cm,y=1.0cm]
\clip(-9.79,-9.12) rectangle (11.88,6.33);
\fill[color=zzttqq,fill=zzttqq,fill opacity=0.1] (1.35,-2.65) -- (4.16,-0.09) -- (1.6,2.72) -- (-1.21,0.16) -- cycle;
\fill[color=zzttqq,fill=zzttqq,fill opacity=0.1] (-2.97,-0.06) -- (-2.04,3.62) -- (1.64,2.69) -- (0.71,-0.99) -- cycle;
\fill[color=zzttqq,fill=zzttqq,fill opacity=0.1] (-3.03,-5.21) -- (0.57,-4.2) -- (-0.40,-0.68) -- (-4.04,-1.6) -- cycle;
\draw [domain=-9.789980199353606:0.0] plot(\x,{(-0--8.97*\x)/-4.99});
\draw [domain=-9.789980199353606:0.0] plot(\x,{(-0-10.49*\x)/-6.28});
\draw [domain=0.0:11.88147788777312] plot(\x,{(-0-0.24*\x)/10.81});
\draw [color=zzttqq] (1.35,-2.65)-- (4.16,-0.09);
\draw [color=zzttqq] (4.16,-0.09)-- (1.6,2.72);
\draw [color=zzttqq] (1.6,2.72)-- (-1.21,0.16);
\draw [color=zzttqq] (-1.21,0.16)-- (1.35,-2.65);
\draw (3,1)-- (3,-1);
\draw (5,1)-- (5,-1);
\draw (-0.54,3.12)-- (-2.3,2.17);
\draw (-1.49,4.88)-- (-3.25,3.93);
\draw (-2.43,-2.02)-- (-0.73,-3.08);
\draw (-3.48,-3.73)-- (-1.78,-4.78);
\draw [color=zzttqq] (-3.03,-5.21)-- (0.57,-4.2);
\draw [color=zzttqq] (0.57,-4.2)-- (-0.40,-0.68);
\draw [color=zzttqq] (-0.40,-0.68)-- (-4.04,-1.6);
\draw [color=zzttqq] (-4.04,-1.6)-- (-3.03,-5.21);

\draw [color=zzttqq] (-2.97,-0.06)-- (-2.04,3.62);
\draw [color=zzttqq] (-2.04,3.62)-- (1.64,2.69);
\draw [color=zzttqq] (1.64,2.69)-- (0.71,-0.99);
\draw [color=zzttqq] (0.71,-0.99)-- (-2.97,-0.06);

\begin{scriptsize}
\fill [color=qqqqff] (0,0) circle (1.5pt);
\draw[color=qqqqff] (0.14,0.22) node {$C$};
\fill [color=qqqqff] (-4.99,8.97) circle (1.5pt);
\draw[color=qqqqff] (-4.22,6.15) node {$D$};
\fill [color=qqqqff] (-6.28,-10.49) circle (1.5pt);
\draw[color=qqqqff] (-4.92,-8.79) node {$A$};
\fill [color=qqqqff] (10.81,-0.24) circle (1.5pt);
\draw[color=qqqqff] (10.95,-0.02) node {$B$};
\fill [color=qqqqff] (1.35,-2.65) circle (1.5pt);
\draw[color=qqqqff] (1.48,-2.34) node {$F$};
\fill [color=qqqqff] (4.16,-0.09) circle (1.5pt);
\draw[color=qqqqff] (4.25,0.14) node {$I$};
\fill [color=uququq] (1.6,2.72) circle (1.5pt);
\draw[color=uququq] (1.69,2.95) node {$J$};
\fill [color=uququq] (-1.21,0.16) circle (1.5pt);
\draw[color=uququq] (-1.09,0.38) node {$K$};
\fill [color=qqqqff] (3,1) circle (1.5pt);
\draw[color=qqqqff] (3.14,1.23) node {$E$};
\fill [color=qqqqff] (3,-1) circle (1.5pt);
\draw[color=qqqqff] (3.14,-0.78) node {$G$};
\fill [color=qqqqff] (5,1) circle (1.5pt);
\draw[color=qqqqff] (5.26,1.23) node {$E_1$};
\fill [color=qqqqff] (5,-1) circle (1.5pt);
\draw[color=qqqqff] (5.26,-0.78) node {$G_1$};
\fill [color=qqqqff] (-0.54,3.12) circle (1.5pt);
\draw[color=qqqqff] (-0.38,3.34) node {$E'$};
\fill [color=qqqqff] (-2.3,2.17) circle (1.5pt);
\draw[color=qqqqff] (-2.13,2.39) node {$G'$};
\fill [color=qqqqff] (-1.49,4.88) circle (1.5pt);
\draw[color=qqqqff] (-1.35,5.11) node {$H$};
\fill [color=qqqqff] (-3.25,3.93) circle (1.5pt);
\draw[color=qqqqff] (-3.13,4.16) node {$L$};
\fill [color=qqqqff] (-2.43,-2.02) circle (1.5pt);
\draw[color=qqqqff] (-2.14,-1.8) node {$E'_1$};
\fill [color=qqqqff] (-0.73,-3.08) circle (1.5pt);
\draw[color=qqqqff] (-0.43,-2.84) node {$G'_1$};
\fill [color=qqqqff] (-3.48,-3.73) circle (1.5pt);
\draw[color=qqqqff] (-3.32,-3.5) node {$M$};
\fill [color=qqqqff] (-1.78,-4.78) circle (1.5pt);
\draw[color=qqqqff] (-1.64,-4.56) node {$N$};
\fill [color=qqqqff] (-3.03,-5.21) circle (1.5pt);
\draw[color=qqqqff] (-2.89,-4.98) node {$O$};
\fill [color=qqqqff] (0.57,-4.2) circle (1.5pt);
\draw[color=qqqqff] (0.72,-3.97) node {$P$};
\fill [color=uququq] (-0.40,-0.68) circle (1.5pt);
\draw[color=uququq] (-0.24,-0.79) node {$Q$};
\fill [color=uququq] (-4.04,-1.6) circle (1.5pt);
\draw[color=uququq] (-3.9,-1.37) node {$R$};
\end{scriptsize}
\end{tikzpicture}
}
\caption{The pattern for covering the junction in the constrained version of the square coverage problem.}
\label{fig:junction_crossing_constrained_sq}
\end{figure}

At the maximum square side length $\ell$ such that the structure
still needs the same number of squares to cover as when using unit
squares, Figure~\ref{fig:junction_crossing_constrained_sq} shows this situation when the junction can
be covered with three squares, among which two squares cover one vertical bar each and
one square covers two vertical bars.
Listing all necessary geometric constraints in Figure~\ref{fig:junction_crossing_constrained_sq} gives
\begin{align}
\label{eq:restricted_start}
\|CK\| &= \sqrt{2}\ell - {\sqrt{2}} / {2} \\
\frac{\|CQ\|}{\sin(45^\circ)} &= \frac{\|CK\|}{\sin(75^\circ)} \\
\|OC\| &= \|OQ\| + \|CQ\| = \|CQ\| + \sqrt{2}\ell \\
\|OC\| &= {5\sqrt{2}}/{4}
\label{eq:restricted_end}
\end{align}

Solving the system of equations in \eqref{eq:restricted_start} to \eqref{eq:restricted_end} gives $\ell\approx 1.289$. 
Considering the possible change of coverage pattern along the odd-length path depicted in Figure~\ref{fig:extreme_box_coverage_pattern}
when $\ell$ reaches $1.25$, we have

\begin{theorem}
    It is NP-hard to approximate the square version of Problem~\ref{p:con_sq_cir_cov} within a factor of $1.25$ when restricting
    the square centers to the region boundary.
\end{theorem}

Compared to the unrestricted version of the square coverage problem, which has an inapproximability ratio of $1.165$, 
the inapproximability ratio for the restricted version of the square coverage problem is higher by around $8.5\%$.

\section{Constant-Factor Approximation Algorithm}
Given the inapprox.\ gaps, it is unlikely that there exists an efficient algorithm that 
runs in polynomial time to approximately solve Problems~\ref{p:cir_cov},~\ref{p:sq_cov},
and~\ref{p:con_sq_cir_cov}. 

In a previous work \cite{Feng2020Optimally}, sampling followed by applying the traditional 
$2$-optimal farthest clustering algorithm \cite{Gonzalez1985Clustering}
for the $k$-center problem gives a $(2+\epsilon)$-optimal result. 
When replacing the farthest clustering algorithm with mathematical programming, 
with the state-of-the-art integer programming
tool \cite{Optimization2019Gurobi}, a $(1+\epsilon)$-optimal result
for $k\leq 100$ can be obtained in $1$ minute on a desktop platform with an Intel 7th Gen Core i7 and 16GB of memory. 

For the square coverage problem, which is similar to the $k$-center problem with the
$L_\infty$ distance metric where the distance between two points $p_1=(x_1,y_1)$ and $p_2=(x_2,y_2)$ 
is $\max(\|x_1 - x_2\|, \|y_1 - y_2\|)$, if we use the $L_\infty$ distance metric, 
which means the square used for covering the target region
is always axis-aligned, using the same farthest clustering algorithm \cite{Gonzalez1985Clustering} can produce a
$2$-optimal solution. 
Besides, a square with side length $\ell$ facing in any direction can be 
fully covered by an axis-aligned square with side length $\sqrt{2}\ell$.
So, squares created with the regular farthest clustering algorithm for the sampled points
using the $L_\infty$ metric give a $(2\sqrt{2} + \epsilon)\approx (2.828 + \epsilon)$-optimal algorithm. 

In the constrained version of the square and circle coverage problems, 
since the farthest clustering algorithm only selects points (circle or square centers) among the
sampled points in the target region or boundary, the results of the previous algorithm automatically 
satisfy the constraints. Thus, the efficient algorithms
for the square and circle coverage problems are still applicable for approximately solving
the constrained circle and square coverage problems within approximation factors of 
$2$ and $2.828$, respectively.

\begin{algorithm}
    \DontPrintSemicolon
    \SetKwInOut{Input}{Input}
	\SetKwInOut{Output}{Output}
	\SetKwInOut{Return}{Return}
	\SetKwComment{Comment}{\%}{}
    \begin{small}
    \KwData{$\mathcal{P}$: a polygon to cover, $k$: the number of squares that can be used to cover the polygon}
    \KwResult{square side length $\ell$, and $k$ squares}
    $S\leftarrow $ Sample $N$ points from the polygon with sampling density $\epsilon$\;
    Run the $2$-optimal farthest clustering algorithm on the sampled points with $L_\infty$ metric to get center points $c_1,\dots, c_k$ and 
    radius $\ell'$, which means $k$ axis-aligned squares centered at $c_1,\dots,c_k$ with side length $\ell = 2\ell' + \epsilon$ can cover the polygon\;
    \Return{$k$ axis-aligned squares, with center points $c_1,\dots,c_k$ and side length $\ell = 2\ell'$.}
    \end{small}
    \caption{Approximate Square Coverage}
\end{algorithm}

\begin{figure}[H]
    \centering
    \includegraphics[width=.49\linewidth, trim={0 0 4cm 0}, clip]{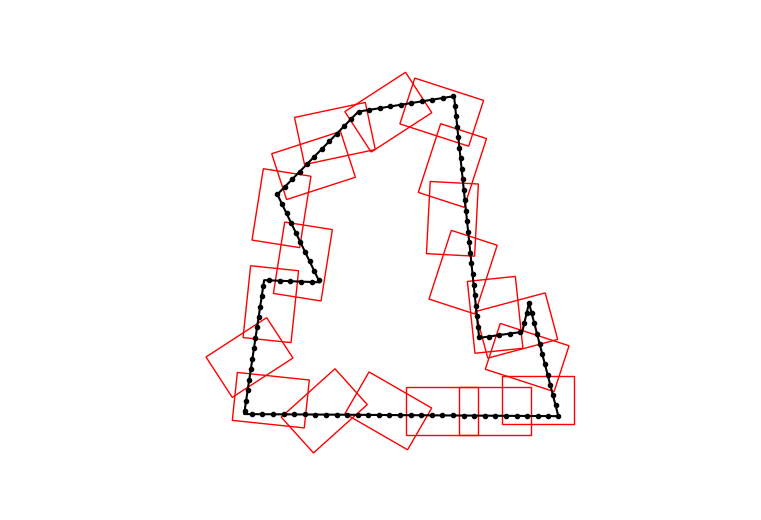}
    \includegraphics[width=.45\linewidth, trim={4cm 0 0 0}, clip]{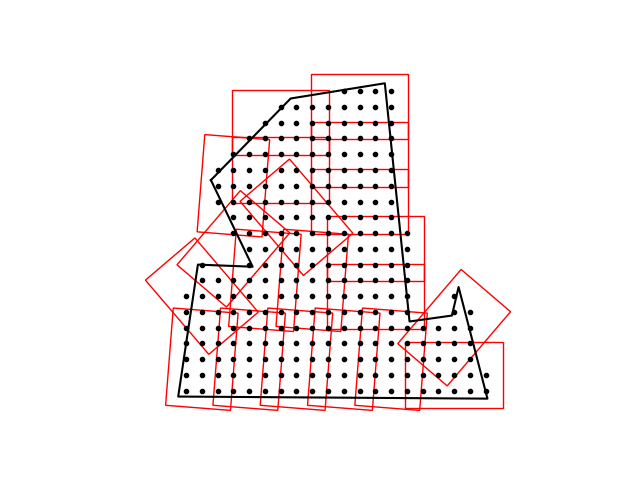}\vspace{-20pt}
    \caption{Examples of results for taking 20 rectangles to cover  
    the perimeter (left) and the interior (right) of a simple polygon using the integer programming based algorithm. 
    Additional padding was added to the results of rectangle side length to the discretized
    initial solution for full coverage of the polygon.}
\end{figure}

For covering a simple polygon with $k$ rectangles, although no polynomial-time approximation algorithm has been developed yet,
the similar idea of sampling followed by applying mathematical programming can be used to provide a coarse result. 
In short, first we discretize the polygon (assuming it is bounded or scaled into a $1\times1$ box) by putting it into a grid with cell size $\epsilon \times \epsilon$
and grid size $N\times N$ (denote $N=1/\epsilon$). 
We then sample $N^3$ possible sensor configurations with $N^2$ location candidates and $N$ candidate orientations for the robot sensing footprint (rectangles)
at each location. Using constraints related to sensing capability between the discretized polygon region and the sensor configurations, we can solve the
problem with widely used mathematical programming tools. 
The implementation for this algorithm adopts OR-Tools \cite{OR-Tools}, which uses SCIP \cite{SCIP} as the integer programming solver.
On a laptop platform with an Intel 11th Gen Core i7 CPU and 16GB of RAM,
covering the perimeter can be solved in 10 seconds for $\epsilon = 1/30$, and
covering the interior can be solved in 1 minute for $\epsilon=1/20$. 

\begin{algorithm}
    \DontPrintSemicolon
    \SetKwInOut{Input}{Input}
	\SetKwInOut{Output}{Output}
	\SetKwInOut{Return}{Return}
	\SetKwComment{Comment}{\%}{}
    \begin{small}
    \KwData{$\mathcal{P}$: a polygon bounded in a $1\times1$ box to cover, $k$: the number of rectangles that can be used to cover the polygon with a given aspect ratio}
    \KwResult{rectangle side length $\ell$, and $k$ rectangles}
    $S\leftarrow $ Sample points from the polygon with a grid of density $1/\epsilon$\;
    $T\leftarrow $ Sample $N^3$ possible sensor configurations with $N^2$ locations and $N$ orientations\;
    Build the integer programming model with constraints related to sensing capability between the discretized polygon region and the sensor configurations\;
    \Return{$k$ rectangles, with center points $c_1,\dots,c_k$ and edge length.}
    \end{small}
    \caption{Approximate Rectangle Coverage}
\end{algorithm}

\section{Conclusions and Future Work}
We studied covering a simple planar polygon with $k$ equal squares or circles of
minimum size, motivated by aerial photography under a limited image budget and
resolution requirements. We proved that approximating the minimum square side length
is NP-hard within a factor of $1.165$, and within a factor of $1.25$ when square
centers are restricted to the region or its boundary. Together with the known
$1.152$-inapproximability for circle coverage, these gaps establish strong hardness
for both unrestricted and location-constrained aerial coverage.

On the algorithmic side, sampling followed by farthest-point clustering under the
$L_\infty$ metric yields a $(2\sqrt{2}+\epsilon)$-approximation for square coverage
and a $(2+\epsilon)$-approximation for circles; both remain valid when centers must
lie inside the region. Closing the still-large gaps between these guarantees and the
corresponding inapproximability bounds, and extending the analysis from squares to
general rectangles, are natural directions for future work.

\section*{Acknowledgements}
The authors thank the anonymous reviewers of AAAI~2026 and WAFR~2026,
as well as Dr.~Yan Lu and Xudong Wang, for their constructive comments
and suggestions, which substantially helped improve the quality of this work.

\bibliographystyle{splncs04}
\bibliography{main}

@article{Hachbaum1985Approximation,
  title={Approximation schemes for covering and packing problems in image processing and VLSI},
  author={Hochbaum, Dorit S and Maass, Wolfgang},
  journal={Journal of the ACM (JACM)},
  volume={32},
  number={1},
  pages={130--136},
  year={1985},
  publisher={ACM New York, NY, USA}
}

@inproceedings{Feder1988Optimal,
  title={Optimal algorithms for approximate clustering},
  author={Feder, Tom{\'a}s and Greene, Daniel},
  booktitle={Proceedings of the twentieth annual ACM symposium on Theory of computing},
  pages={434--444},
  year={1988}
}

@inproceedings{Roberts2017Submodular,
  title={Submodular trajectory optimization for aerial 3d scanning},
  author={Roberts, Mike and Dey, Debadeepta and Truong, Anh and Sinha, Sudipta and Shah, Shital and Kapoor, Ashish and Hanrahan, Pat and Joshi, Neel},
  booktitle={Proceedings of the IEEE International Conference on Computer Vision},
  pages={5324--5333},
  year={2017}
}

@article{Meurer2017Sympy,
  title={SymPy: symbolic computing in Python},
  author={Meurer, Aaron and Smith, Christopher P and Paprocki, Mateusz and {\v{C}}ert{\'\i}k, Ond{\v{r}}ej and Kirpichev, Sergey B and Rocklin, Matthew and Kumar, AMiT and Ivanov, Sergiu and Moore, Jason K and Singh, Sartaj and others},
  journal={PeerJ Computer Science},
  volume={3},
  pages={e103},
  year={2017},
  publisher={PeerJ Inc.}
}

@article{Mohar2001Face,
  title={Face covers and the genus problem for apex graphs},
  author={Mohar, Bojan},
  journal={Journal of Combinatorial Theory, Series B},
  volume={82},
  number={1},
  pages={102--117},
  year={2001},
  publisher={Elsevier}
}

@inproceedings{Feng2019Efficient,
  title={Efficient algorithms for optimal perimeter guarding},
  author={Feng, Si Wei and Han, Shuai D and Gao, Kai and Yu, Jingjin},
  booktitle={2019 Robotics: Science and Systems},
  year={2019}
}

@inproceedings{Feng2020Optimally,
  title={Optimally Guarding Perimeters and Regions with Mobile Range Sensors},
  author={Feng, Si Wei and Yu, Jingjin},
  booktitle={2020 Robotics: Science and Systems},
  year={2020}
}

@inproceedings{Feng2021Sensor,
  title={Sensor placement for globally optimal coverage of 3d-embedded surfaces},
  author={Feng, Si Wei and Gao, Kai and Gong, Jie and Yu, Jingjin},
  booktitle={2021 IEEE International Conference on Robotics and Automation (ICRA)},
  pages={8600--8606},
  year={2021},
  organization={IEEE}
}

@article{Gage1992Command,
  title={Command control for many-robot systems},
  author={Gage, Douglas W},
  journal={Unmanned Systems Magazine},
  volume={10},
  number={4},
  pages={28--34},
  year={1992}
}

@article{Optimization2019Gurobi,
  title={Gurobi Optimizer 9.0},
  author={Gurobi Optimization},
  journal={Gurobi: http://www.gurobi.com},
  year={2019}
}

@article{Gonzalez1985Clustering,
  title={Clustering to minimize the maximum intercluster distance},
  author={Gonzalez, Teofilo F},
  journal={Theoretical computer science},
  volume={38},
  pages={293--306},
  year={1985},
  publisher={Elsevier}
}

@book{toth2017handbook,
  title={Handbook of discrete and computational geometry},
  author={Toth, Csaba D and O'Rourke, Joseph and Goodman, Jacob E},
  year={2017},
  publisher={CRC press}
}

@misc{OR-Tools,
  title={OR-Tools: Google's Optimization Tools},
  author={Google},
  url={https://developers.google.com/optimization},
  year={2024}
}

@article{SCIP,
  title={SCIP: Solving Constraint Integer Programs},
  author={Achterberg, Tobias and Berthold, Timo and Koch, Thorsten and Pfetsch, Marc E and Wolter, Kathrin},
  journal={Mathematical Programming Computation},
  volume={11},
  number={1},
  pages={1--41},
  year={2019},
  publisher={Springer}
}

\end{document}